\documentclass[journal,comsoc]{IEEEtran}
\IEEEoverridecommandlockouts
\usepackage{cite}
\usepackage{amsmath,amssymb,amsfonts}
\usepackage{algorithmic}
\usepackage{graphicx}
\usepackage{textcomp}
\usepackage{xcolor}
\usepackage{float}
\usepackage{amsmath,amssymb}
\usepackage{bm}
\usepackage{bbm}
\usepackage{multirow} 
\usepackage{caption}
\usepackage[ruled,vlined]{algorithm2e}
\usepackage{booktabs}
\usepackage{array}
\usepackage{subfig}
\usepackage{soul}
\usepackage{rotating}

\newcommand{\vect}[1]{\mathbf{#1}}
\newcommand{\matr}[1]{\mathbf{#1}}

\newcommand{\set}[1]{\mathcal{#1}}

\def\BibTeX{{\rm B\kern-.05em{\sc i\kern-.025em b}\kern-.08em
		T\kern-.1667em\lower.7ex\hbox{E}\kern-.125emX}}
\usepackage[T1]{fontenc}

\begin{document}

\title{COFT-AD: COntrastive Fine-Tuning for Few-Shot Anomaly Detection}

\author{Jingyi~Liao,
        Xun~Xu,~\IEEEmembership{Senior Member,~IEEE,}
        Manh~Cuong~Nguyen,
        Adam Goodge,
        and~Chuan~Sheng~Foo
\thanks{All authors are with the Institute for Infocomm Research (I2R), Agency for Science, Technology and Research (A*STAR). Chuan Sheng Foo is also with the Centre for Frontier AI Research (CFAR), Agency for Science, Technology and Research (A*STAR).
Correspondence to Xun~Xu e-mail: alex.xun.xu@gmail.com.}%
}

\maketitle
\begin{abstract}

Existing approaches towards anomaly detection~(AD) often rely on a substantial amount of anomaly-free data to train representation and density models. However, large anomaly-free datasets may not always be available before the inference stage; in which case an anomaly detection model must be trained with only a handful of normal samples, a.k.a. few-shot anomaly detection (FSAD). In this paper, we propose a novel methodology to address the challenge of FSAD which incorporates two important techniques. Firstly, we employ a model pre-trained on a large source dataset to initialize model weights. Secondly, to ameliorate the covariate shift between source and target domains, we adopt contrastive training to fine-tune on the few-shot target domain data. To learn suitable representations for the downstream AD task, we additionally incorporate cross-instance positive pairs to encourage a tight cluster of the normal samples, and negative pairs for better separation between normal and synthesized negative samples. We evaluate few-shot anomaly detection on on 3 controlled AD tasks and 4 real-world AD tasks to demonstrate the effectiveness of the proposed method.

\end{abstract}


\IEEEpeerreviewmaketitle

\section{Introduction}

Identifying anomalous data, a.k.a. anomaly detection~(AD), is a fundamental problem in machine learning research with significant potential in various applications, such as defect identification~\cite{bergmann2019mvtec}, surveillance~\cite{xiang2008video}, and autonomous driving~\cite{lis2019detecting}.
It is typically assumed that normal images are abundant, whereas labeled  anomaly examples are non-exhaustive. Hence, anomaly detection models are trained only with normal images, a.k.a. one-class classification. However, this is not always the case. For example, some practical applications require an AD system to function instantly upon deployment on a new task with limited existing data or when acquiring normal samples is expensive or time-consuming. As a result, large numbers of normal images may not be available for building a good anomaly detection model, especially at the initial stages of bootstrapping an AD system.

In response to the demand for building an anomaly detection system under the deficiency of normal training examples, few-shot anomaly detection~(FSAD), where only a few normal and no abnormal images are available, emerged as a solution~\cite{sheynin2021hierarchical,huang2022registration,graphcore2023}, and there is growing interest from the community on this topic.
Among the limited pioneering works, Sheynin et al.\cite{sheynin2021hierarchical} developed a generative adversarial model to distinguish transformed image patches from generated ones. However, such adversarial models may be tricky to tune\cite{kodali2017convergence} and require multiple transformations on test samples at inference time, resulting in additional computation overhead. The more recent work of RegAD~\cite{huang2022registration} learns a common model over multiple classes of normal images using a feature registration proxy task, but their method requires a training set with normal images from multiple known classes, which is a more restrictive setting. GraphCore~\cite{graphcore2023}, stemming from PatchCore~\cite{roth2022towards}, approached FSAD by comparing testing patches to few-shot training ones. However, both RegAD and GraphCore are built upon the assumption that anomalies are caused by variations in local appearance and {their success is only demonstrated } on highly constrained industrial anomaly detection tasks. In summary, existing methods in FSAD either complicate the training procedure with tricky hyper-parameter tuning~\cite{sheynin2021hierarchical} or have overly restrictive assumptions on the presentations of anomalies~\cite{roth2022towards,graphcore2023}, which limit their applicability to {wider range of } tasks.

\begin{figure}
    \centering
    \includegraphics[width=1\linewidth]{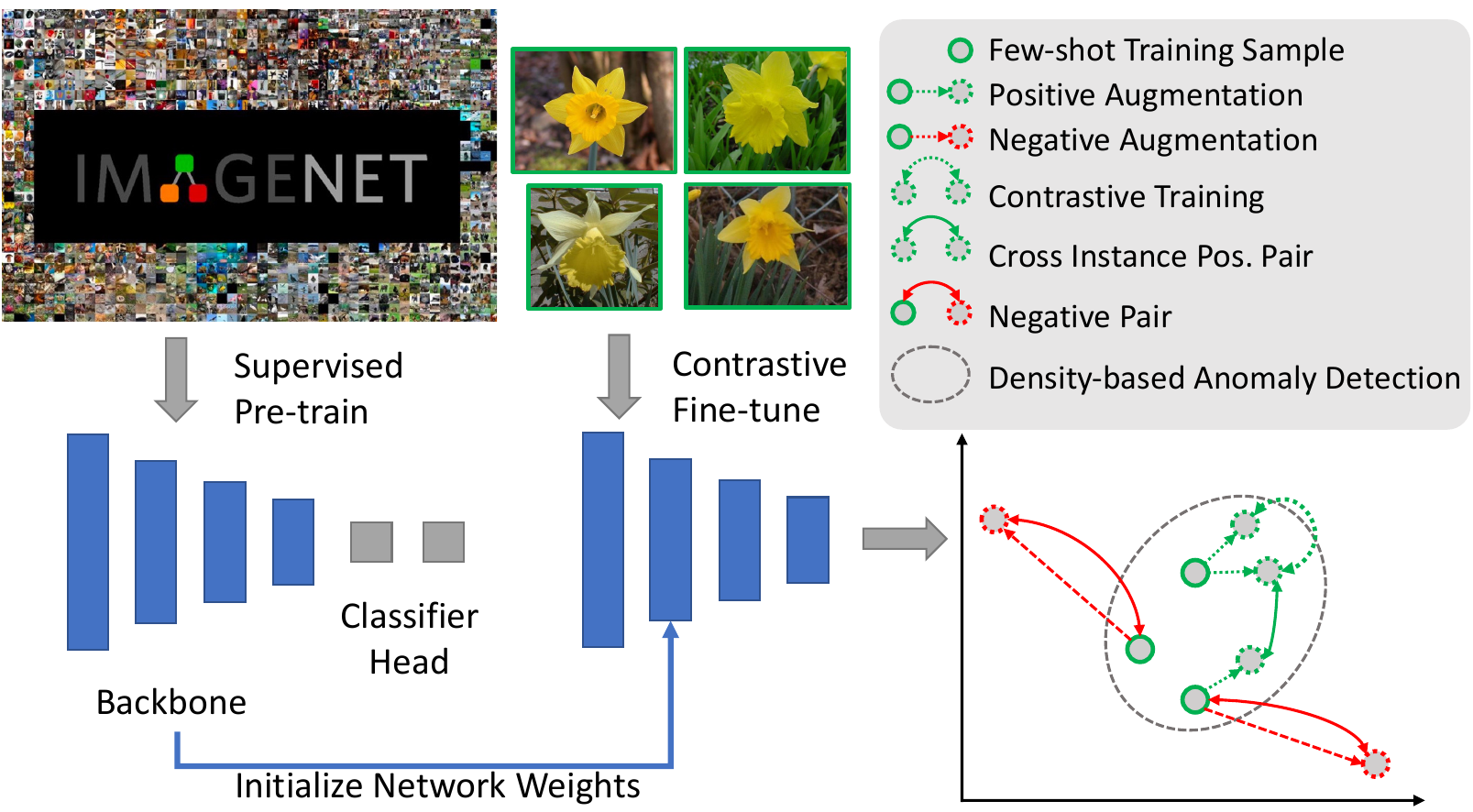}
    \caption{We present a contrastive fine-tuning approach towards few-shot anomaly detection. Backbone network is initialized by ImageNet supervised pre-trained model weights. Fine-tuning on few-shot target dataset through contrastive training achieves few-shot anomaly detection.}
    \label{fig:thumbnail}
\end{figure}

In this work, we aim to develop a few-shot anomaly detection framework applicable to a wide variety of data domains, without complicated hyper-parameter tuning. We achieve this by synergistically combining transfer learning from a pre-trained model with representation learning on the available few-shot normal data. 
Re-using weights from a backbone network pre-trained on a large source domain dataset, e.g. ImageNet~\cite{russakovsky2015imagenet}, enables {exploitation of } powerful, low-level feature extractors and better initialization of network parameters~\cite{kornblith2019better} for downstream tasks. 
We believe that re-using pre-trained weights could particularly benefit few-shot anomaly detection, {as training data is insufficient to achieve learning} good representations from scratch. However, as highlighted in existing works~\cite{xu2022revisiting,li2021unsupervised}, directly reusing the pre-trained weights may not fully utilize the limited supervision in the downstream task because of two factors. First, the anomaly detection task requires feature representation that separates normal samples from abnormal ones. The representations learned during pre-training are {optimized} for a different task, such as semantic image classification of ImageNet, and are therefore not necessarily optimal for the downstream anomaly detection task. Second, performance can be hindered by covariate shift~\cite{wang2018deep}, which is a result of the difference in distribution between the source domain data~(used for pre-training the model), and the target domain data~(anomalies are to be detected).

To tackle these challenges, we propose an unsupervised fine-tuning approach, named ``COntrastive Fine-Tuning for few-shot Anomaly Detection''~(COFT-AD), to adapt pre-trained weights to downstream few-shot anomaly detection task.
To ameliorate the covariate shift between source and target domain data, we first design customized loss functions with three components, each of which encourages a desirable property in the learnt representation space when adapted towards the anomaly detection task in the target domain.
The first component is a contrastive loss~\cite{grill2020bootstrap}, defined on all available few-shot normal examples. Optimizing a contrastive loss helps adapt pre-trained model weights to downstream anomaly detection dataset distribution. Specifically, this is achieved by encouraging closeness between one sample and its augmented versions in the feature space. This helps improve model's robustness to natural variations within the normal data.
The second component further optimizes representations for the downstream anomaly detection task by encouraging normal samples to form a tight cluster in the feature space such that a density based anomaly inference is facilitated. We achieve this with a cross-instance positive pair loss that encourages closeness between the feature embeddings of two randomly sampled normal images. Note that this differs from standard contrastive training as closeness is encouraged between two \emph{different} normal samples instead of a sample and its augmented version. 
Finally, when prior knowledge on the anomalies is available, we are able to synthesize negative examples~\cite{li2021cutpaste} to simulate anomalies in an additional negative pair loss, which encourages better separation between the normal samples and their negative pairs. In our analysis, we empirically reveal that the method, by which negative samples are synthesised, has high impact on FSAD performance and should be used carefully with prior knowledge on the nature of anomalies. An overall illustration of the proposed framework is presented in Fig.~\ref{fig:thumbnail}.

We extensively evaluate COFT-AD on a diverse range of anomaly detection tasks, covering controlled setups, in which normal and anomalous data are deliberately designed in order to exhibit certain properties, as well as on real-world data FSAD, where the nature of anomalies is out of our control. The extensive evaluations demonstrate the ability of our method to handle greater diversity in data domains while maintaining its strong performance. We summarize the contribution of our work as follows:

\begin{itemize}
    \item We approach hyper-parameter friendly and data domain versatile few-shot anomaly detection from a transfer learning perspective. We adopt contrastive fine-tuning on few-shot normal samples in the target domain to adapt pre-trained model weights to target domain distribution.
    \item We further introduce a cross-instance positive pair loss to encourage normal samples to form a tight cluster in the embedding space for better density-based anomaly detection. 
    \item In scenarios where prior knowledge on anomaly presentations is available, a negative pair loss is further incorporated with synthesized negative samples to encourage better separation between normal and abnormal samples.
    \item We demonstrate the effectiveness of the proposed method, COntrastive Fine-Tuning for few-shot Anomaly Detection~(COFT-AD), on 3 controlled anomaly detection datasets and 4 real-world industrial defect identification datasets, achieving very competitive performance.
\end{itemize}

\begin{figure*}[!htb]
    \centering
    \includegraphics[width=0.95\linewidth]{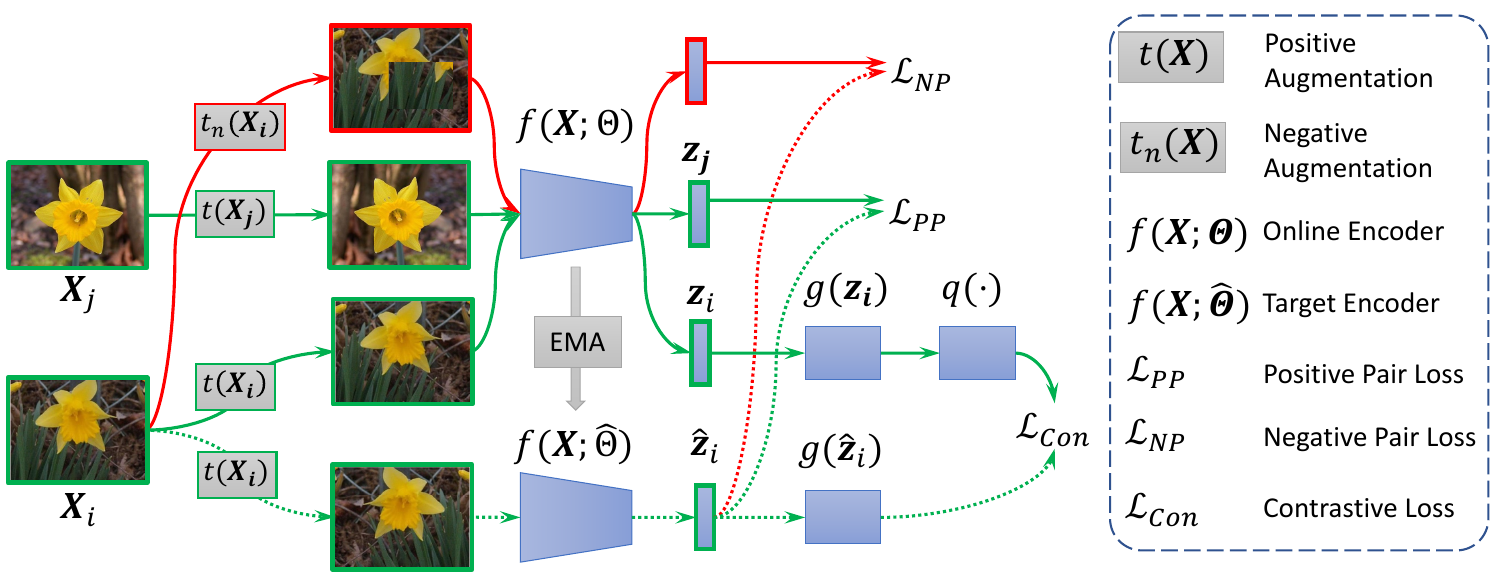}
    \caption{Illustration of adapting source domain pretrained model through combining contrastive training loss (green arrow lines), cross-instance positive pair loss~(green arrow lines) and negative pair loss~(red arrow lines) for few-shot anomaly detection. The dashed arrow lines indicate no gradient backpropagation.}
    \label{fig:network}
\end{figure*}

\section{Related Work}
\subsection{Anomaly Detection}

Traditional AD methods include PCA \cite{shyu2003novel}, cluster analysis \cite{kim2012robust} and one-class classification \cite{oc_svm}. With the advent of deep learning, representation learning is employed to avoid manual feature engineering and kernel construction. This leads to novel anomaly detection methods based on deep neural networks, such as autoencoders \cite{AE_SSIM} and generative adversarial networks (GAN) \cite{oc_gan, anogan}. Among them, anoGAN \cite{anogan} was proposed to learn the manifold of normal samples and anomalous samples cannot be perfectly projected onto the normal manifold by the generator learned solely with normal samples. However, it requires expensive optimization for detecting abnormal samples and training GANs is prone to some well-known challenges including instability and mode collapse. Among the autoencoder based approaches, \cite{AE_SSIM} adopted the SSIM metric as the similarity measure between input and reconstructed images. 
Recently, an effective line of works approach AD as a two-stage problem of representation learning followed by outlier detection in the learned representation space~\cite{ruff2018deep,golan2018deep,yi2020patch,sohn2021learning,8721681}. Among these works, deep SVDD~\cite{ruff2018deep} proposes to learn a feature embedding that groups normal samples closer to a cluster center. Subsequent works develop self-supervised pretraining methods to learn representations suitable for separating abnormal samples from normal ones by optimizing a proxy task~\cite{golan2018deep,yi2020patch,sohn2021learning,li2021cutpaste}. Anomaly detection is then achieved through a density model fit to the learnt representations of normal training samples. These approaches prevail in many anomaly detection benchmarks and are computationally efficient. Nevertheless, representation learning requires a substantial amount of training data which may not be readily available in certain practical applications. Another line of works approach anomaly detection through a non-parametric framework. PatchCore~\cite{roth2022towards} divides images into multiple patches and a nearest neighbour classifier is adopted to compare testing patches to training ones to determine anomalies. To reduce computational cost, only a subset of training patches are selected for comparison. {Nonetheless}, the non-parametric approaches still suffer from computational efficiency when the diversity of normal pattern grows larger.
Recently, an emerging topic in anomaly detection tackles normal data distribution shift in testing data. Specifically, \cite{cao2023anomaly} is built upon reverse distillation~\cite{deng2022anomaly} for anomaly detection. It proposed to improve the generalization of anomaly detection model by augmenting testing data with selected data augmentation.

\subsection{Few-Shot Anomaly Detection} 
Few-shot anomaly detection~(FSAD) is an emerging problem which aims to enable anomaly detection based on only a few normal samples as training data. We distinguish FSAD from the few-shot semi-supervised anomaly detection setting in~\cite{ruff2019deep}, where a limited number of \emph{labeled anomalies} are available for training, as it is sometimes referred to as few-shot anomaly detection in the literature~\cite{pang2021explainable}.
The pioneering FSAD method of \cite{sheynin2021hierarchical} employs a hierarchical generative model to generate new samples from the few-shot examples. A discriminator is designed to discriminate generated images from the real ones and different transformations. Anomalies are then determined by whether the discriminator can correctly classify the type of transformations. Graphcore~\cite{graphcore2023} extends Patchcore~\cite{roth2022towards} for few-shot anomaly detection by focusing on patch level anomalies.Inspired by the few-shot learning paradigm, RegAD~\cite{huang2022registration} develops a registration-based proxy task for representation learning; this task aims to find the affine transformation that aligns the feature maps of two samples from the same semantic class. 
RegAD requires additional related training data, for instance, data from other classes besides the target class on the MVTec dataset, for training the proxy task. The work of~\cite{andoanomaly} addresses a different few-shot setting that requires normal samples to be provided with semantic labels. When normal data comprise  multiple semantic classes, embedding all normal samples into a single cluster may result in a failure to detect anomalies occurring between semantic classes. Learning multiple prototypes was proposed to tackle this issue. In comparison, our method adapts pre-trained weights to target data using only a few normal training samples; unlike some existing methods, no additional data are required during the representation learning phase. This enables our method to perform well in a broader range of anomaly detection scenarios. Recently, an emerging topic in anomaly detection tackles normal data distribution shift in testing data. Specifically, \cite{cao2023anomaly} is built upon reverse distillation~\cite{deng2022anomaly} for anomaly detection. It proposed to improve the generalization of anomaly detection model by augmenting testing data with selected data augmentation. Few-shot anomaly detection provides a new pespective for tackling the distribution shift issue.

\subsection{Contrastive Learning}
Pre-training feature representation through contrasting augmented samples of the same identity has demonstrated promising results. SimCLR~\cite{chen2020simple,he2020momentum} employed an N-pair loss~\cite{sohn2016improved} to encourage two augmentations of the same instance~(positive pairs) to be close in the feature space and other instances~(negative pairs) to be far away. The use of negative pairs requires a large training batch size, BYOL~\cite{grill2020bootstrap} introduced an exponential moving average model to avoid collapsed predictions and remove negative pairs. Besides pre-training of representations, contrastive learning has been recently demonstrated to be effective for label-efficient fine-tuning~\cite{liu2021ttt++,xu2022revisiting,chen2022contrastive,li2021unsupervised}. When source and target domain data distributions are subject to covariate shift, contrastive training on the target data in an unsupervised fashion can potentially alleviate the domain shift~\cite{xu2022revisiting,li2021unsupervised}. In the context of anomaly detection, contrastive training has been adopted for various purposes.

CSI~\cite{tack2020csi} uses negative pairs in a different way, they consider different samples within the same category as negative pairs and does not have any positive pairs, whereas our method considers them as positive pairs and uses synthesized anomaly samples as negative pairs. In addition, CSI does not use a pretrained model, making it not optimal for few-shot scenarios. Moreover, the loss design in CSI differs from ours. While both methods build on contrastive loss, CSI incorporates an additional transformation prediction loss.
Contrastive learning has been employed for separating normal samples from abnormal ones in the feature {space}~\cite{9913887}. In \cite{9913887}, as negative pairs are constructed between different instances, a carefully selection of negative pairs is vital to the performance of anomaly detection, and this could be sensitive to data samples with low intra-class variations. Other self-supervised learning approaches use auxiliary tasks for representation learning for anomaly detection~\cite{10012327}.
In this work, we demonstrate that contrastive training in conjunction with positive pair and negative pair training on the target domain data plays an important role in adapting pre-trained model weights to the target domain distribution, which helps anomaly detection with limited exposure to positive or normal samples.

\section{Methodology}

In this work, we assume a model pre-trained on a large external image collection (e.g. ImageNet) is available. We refer to this external data as the source domain. The downstream data where we want to detect anomalies is referred to as the target domain. We first describe contrastive training as a means to fine-tune a pre-trained model to the target domain distribution. We then introduce the cross-instance positive pair loss to encourage normal samples to form a cluster in the feature space, and the negative pair loss to encourage better separation of normal and abnormal samples when prior knowledge of anomalies is available. An overview of the proposed adaptation framework is shown in Fig.~\ref{fig:network}. Lastly, we describe how to build the density-based anomaly detection model on the learnt representations.

\subsection{Contrastive Training for Fine-tuning}
We first denote the few-shot training examples from the target domain as $\set{D}_{T}=\{\matr{X}_i\}_{i=1\cdots N_T}$.
The backbone, pre-trained network $f$ is parameterized by $\matr{\Theta}$, and $\vect{z}=f(\matr{X};\matr{\Theta})$ are the embeddings of input samples $\matr{X}$ in the feature space.  Contrastive training updates the model parameters $\matr{\Theta}$ by optimizing a contrastive loss in an unsupervised manner as in Eq.~\ref{eq:BYOL}. In this work, we consider the BYOL~\cite{grill2020bootstrap} method for contrastive learning, which uses a duo of online and target encoder networks, due to its smaller memory requirements.

\begin{equation}\label{eq:BYOL}
    \mathcal{L}_{Con}=-\frac{1}{N_T}\sum_{\matr{X}_i\in\set{D}_T}\frac{q(g(\matr{z}_i))^\top {g(\hat{\matr{z}}_i)}}{||q(g({\matr{z}_i}))||\cdot||{g(\hat{\matr{z}}_i})||}
\end{equation}

To learn effective representations, contrastive training contrasts between two random augmentations of the same input image, denoted as $t(\matr{X})$. The encoder network outputs the representation embedding for each augmented input as $\matr{z}=f(t(\matr{X});\Theta)$. The representations are further projected to a lower dimensional space through a projection head $g(\cdot)$. The cosine similarity is then calculated between the predictor's output $q(g(\matr{z}))$ on the online view and the projector's output $g(\hat{\matr{z}})$ on the target view. To avoid a trivial solution, e.g. an encoder function giving constant outputs, the target view is the output of an exponential moving average model, i.e. $\hat{\matr{z}}=f(\matr{X};\hat{\Theta})$ and $\hat{\Theta}_{t}=\beta\hat{\Theta}_{t-1}+(1-\beta)\Theta_t$ where $\hat{\Theta}$ are the parameters of the target network and $\beta$ is a moving average hyper-parameter. When a source domain model with trained parameters $\Theta^S$ is available, contrastive training on the target domain is initialized by this source domain model, i.e. $\Theta_0=\Theta^S$, such that the low-level feature extractors can be reused. Therefore, contrastive training serves to adapt a pre-trained network to the few-shot target domain training samples. 

\noindent\textbf{Discussion}: \label{sec:app_contrast}
Recent works have demonstrated that contrastive training helps adapt model parameters to target domain distributions~\cite{liu2021ttt++,xu2022revisiting,chen2022contrastive,li2021unsupervised}. We argue that contrastive training on the target domain data can alleviate the negative impact of covariate shift. 
For simplicity, we denote the source domain dataset as $\set{D}_{S}$, e.g. the ImageNet dataset, and the target domain dataset as $\set{D}_{T}$, e.g. anomaly detection dataset. The objective of supervised training on the source domain is to find the optimal model parameters $\Phi^{S*},\Theta^{S*}$ that minimize the following cross entropy loss where $h(\cdot;\matr{\Phi})$ is the classifier on the source domain and $f(\cdot;\matr{\Theta})$ is the backbone network to be transferred:
\begin{equation}
    \Phi^{S*},\Theta^{S*} = \arg\min_{\Phi,\Theta} \frac{1}{|\set{D}_{S}|}\sum_{{\matr{X}_i,y_i\in \set{D}_{S}}}\mathcal{L}_{CE}(h(f(\matr{X}_i;\Theta);\Phi),y_i)\\
\end{equation}

Covariate shift between source and target dataset indicates a distributional misalignment, i.e. $p_{S}(\matr{X})\neq p_{T}(\matr{X})$ which is easily manifested by the difference in the contents of the source and target domain data. Therefore, it is reasonable to believe the backbone network optimized for source domain model is not optimal for the target domain distribution. To ease the negative impact caused by covariate shift, we introduce contrastive training on the target domain by optimizing an unsupervised contrastive loss with model parameters initialized with the source domain, as in Eq.~\ref{eq:ContrastiveFT}.
\begin{equation}\label{eq:ContrastiveFT}
\resizebox{0.91\linewidth}{!}{
$
\begin{split}
    \min_{\Theta}\frac{1}{|\set{D}_T|}&\sum_{\matr{X}_i\in\set{D}_T}\mathcal{L}_{Con}(f(t(\matr{X}_i);\Theta),f(t(\matr{X}_i);\hat{\Theta})),\quad \\
    &s.t. \; \Theta^T_0 = \Theta^{S*}
\end{split}
$
}
\end{equation}

By minimizing the contrastive loss, the network is able to capture key features from the target domain to discriminate non-identical instances and we empirically demonstrate this to be effective for adapting our source model to the target domain for downstream anomaly detection. When the augmentations are chosen to mimic the commonly seen variations within normal samples, contrasting two augmented images forces the network to produce similar representations regardless of the augmentations. This means the representation learned from contrastive training allows the network to learn features invariant to common variations in appearance and pose that one could encounter within the normal data. Such ability will help bring normal samples closer in the feature space, thus benefit downstream anomaly detection.

\subsection{Cross-Instance Positive Pair Loss}

Detecting anomaly with pre-trained representation can be often formulated as fitting a parameterized distribution model, e.g. a multi-variate Gaussian distribution, to the normal samples~\cite{ruff2018deep}. Such a density-based anomaly detection paradigm introduces low storage overhead and efficient computation of anomaly score.
The contrastive training objective introduced in the previous section encourages adaptation to target distribution, but it does not guarantee the learned representation is suitable for the downstream density-based anomaly detection, in which normal data is assumed to occupy tight clusters while anomalies are sparse. 
Inspired by the success of one-class classification~\cite{ruff2018deep}, we propose to encourage normal samples to form a tight cluster in the feature space through an additional cross-instance positive pair loss term. Specifically, we treat a pair of randomly selected normal samples as a positive pair, the representations of each positive pair are encouraged to be closer by minimizing the cosine similarity as in Eq.~\ref{eq:PosPair} where $\vect{p}$ is a random permutation of the list $\{1,\cdots, N\}$. 

\begin{equation}\label{eq:PosPair}
\resizebox{0.8\linewidth}{!}{
$
\begin{split}
\mathcal{L}_{PP}= &-\frac{1}{2N_T} \sum\limits_i \sum\limits_{j\in\vect{p}} \frac{f(t(\matr{X}_i);\Theta)^\top f(t(\matr{X}_j);\hat{\Theta})}{||f(t(\matr{X}_i);\Theta)||\cdot ||f(t(\matr{X}_j);\hat{\Theta})||} \\&+ \frac{f(t(\matr{X}_j);\Theta)^\top f(t(\matr{X}_i);\hat{\Theta})}{||f(t(\matr{X}_j);\Theta)||\cdot ||f(t(\matr{X}_i);\hat{\Theta})||}
\end{split}
$
}
\end{equation}

Compared with the alternative of maintaining a fixed cluster center as proposed in \cite{ruff2018deep}, the cross-instance positive pair loss has two advantages. First, we do not need to fix the cluster center at the start of  training. This avoids introducing excessive regularization on the representation embedding as the cluster center may vary during the course of training. Second, we minimize the cosine similarity between the online view and target view where the latter does not backpropagate gradients. This avoids collapse to a trivial solution (e.g. all zero weights).
We note that the cross-instance positive pair loss is calculated on the features directly from the backbone network. This is due to the fact that backbone output features will be used for anomaly detection so the loss should be optimized in the feature space.

\subsection{Incorporating Negative Pair Loss}

The contrastive loss and cross-instance positive pair loss both encourage similarity in embeddings of data within the normal distribution. However, we also want to ensure a significant distance in the feature space exists between the normal samples and anomalies. Synthesized negative examples have been demonstrated to aid in pre-training representations for anomaly detection. Well-calibrated synthesis approaches, where the negative samples closely align with the real anomalies, can even achieve the state-of-the-art performance on certain datasets~\cite{li2021cutpaste}. In this work, we propose to incorporate additional synthetic negative examples when prior knowledge about the nature of anomalies is available. 
We define prior knowledge as an understanding of the types of anomalies that are likely to occur. This knowledge can be acquired through previous experiences. For example, in semiconductor packaging quality control, where products are scanned by a fixed microscope, images are always acquired in the same pose. In such a circumstance, geometric transformations will not simulate potential anomalies. While the potential anomalies may arise from the voids in the solder which visually look like a circular hole in the acquired image and we could exploit this prior knowledge to synthesize negative pairs for adapting model for anomaly detection.

Specifically, denoting a synthesized negative sample as $t_n(\matr{X})$, to encourage better separation between normal and abnormal samples we minimize the cosine similarity between the original image embedding and its negative pair embedding as below: 
\begin{equation}
    \mathcal{L}_{NP}=\frac{1}{N_T}\sum_i\frac{f(\matr{X}_i;\hat{\Theta})^\top f(t_n(\matr{X}_i);{\Theta})}{||f(\matr{X}_i;\hat{\Theta})||\cdot||f(t_n(\matr{X}_i);{\Theta})||}
\end{equation}

It is worth noting that the negative contrasting is also carried out directly on the backbone output features to reflect the constraints applied to the feature representations. A related design was presented in \cite{ruff2019deep} for semi-supervised anomaly detection by minimizing the reciprocal of the distance between annotated anomalies and normal sample cluster center. Again, we believe minimizing the cosine similarity is compatible with the contrastive training objective, which also optimizes the cosine similarity, and cross-instance positive pair loss has no risk of having a trivial solution. To differentiate the augmentations employed, we referred to $t(\matr{X})$ as \textbf{positive augmentation} and $t_n(\matr{X})$ as \textbf{negative augmentation}.

The final training loss is a weighted sum of the three loss terms:
\begin{equation}
    \mathcal{L}_{all}=\mathcal{L}_{Con} + \lambda_{PP}\mathcal{L}_{PP} + \lambda_{NP}\mathcal{L}_{NP}
\end{equation}

\subsection{Density-based Anomaly Detection}

To perform anomaly detection using the learnt representations, we follow the density-based approach in~\cite{li2021cutpaste} and fit a multivariate Gaussian distribution to the few-shot normal samples. Note that the learnt feature representations must be L2-normalized before density estimation and inference because during representation learning, we optimize the cosine similarity which is agnostic to the magnitude of feature representations. Moreover, to increase the amount of data for fitting the Gaussian distribution we produce $N_A$ times augmented samples from the few-shot normal samples. Formally, the mean $\mu$ and covariance $\Sigma$ is obtained through maximum likelihood estimation as below where 
$\set{D}_{TA}=\underbrace{\set{D}_T\cup\set{D}_T\cup\cdots\set{D}_T\,}_\text{$N_A$ times}$.


\begin{equation}
\resizebox{0.9\linewidth}{!}{
$
\begin{split}
    &\vect{\mu} = \frac{1}{|\set{D}_{TA}|}\sum\limits_{\matr{X}_i\in\set{D}_{TA}}\frac{f(t(\matr{X}_i))}{||f(t(\matr{X}_i))||},\;\\
    &\matr{\Sigma}=\frac{1}{|\set{D}_{TA}|}\sum\limits_{\matr{X}_i\in\set{D}_{TA}}\left(\frac{f(t(\matr{X}_i))}{||f(t(\matr{X}_i))||}-\vect{\mu}\right)\left(\frac{f(t(\matr{X}_i))}{||f(t(\matr{X}_i))||}-\vect{\mu}\right)^\top
\end{split}
$
}
\end{equation}

The anomaly score is then given by the Mahalanobis distance as in Eq.~\ref{eq:MahaDist} and test samples are ranked by the anomaly score for anomaly detection. We add a small epsilon to the diagonal entries of $\Sigma$ to ensure that it is invertible.

\begin{equation}\label{eq:MahaDist}
    d_{AS}(\matr{X}) =\sqrt{\left(\frac{f(\matr{X})}{||f(\matr{X})||}-\vect{\mu}\right)^\top\matr{\Sigma}^{-1}\left(\frac{f(\matr{X})}{||f(\matr{X})||}-\vect{\mu}\right)}
\end{equation}

\section{Experiment}
We evaluate the performance of COFT-AD on three controlled anomaly detection datasets and four industrial defect identification datasets. We benchmark against state-of-the-art anomaly detection methods and achieve very competitive performance. We also carry out ablation studies on individual components of our methodology and provide further analysis of its behaviours.

\begin{figure*}[!htb]
    \centering
    \subfloat[Flowers17 dataset]{
    \includegraphics[width=1\linewidth]{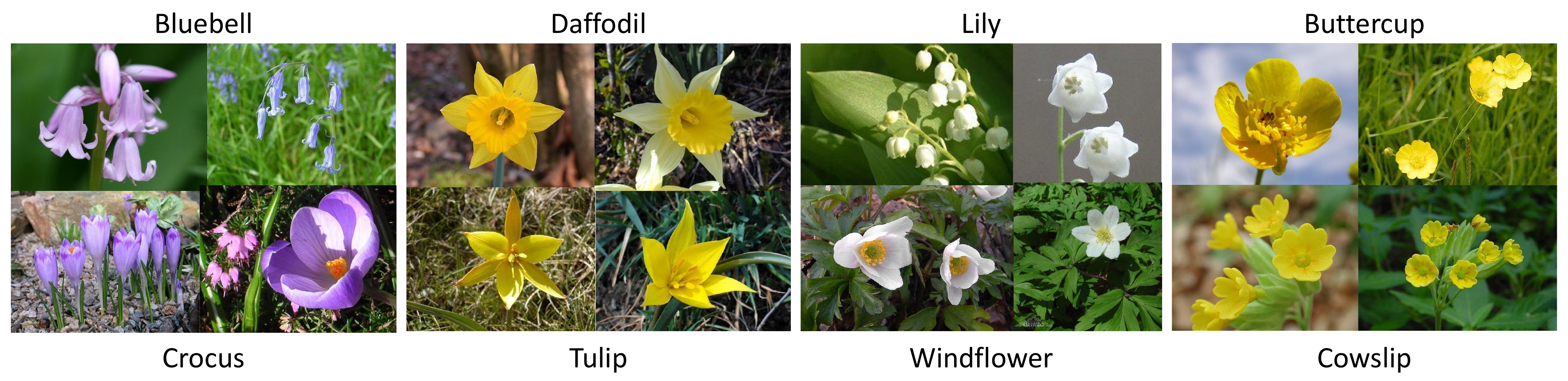}}\\
    \vspace{-0.4cm}
    \subfloat[Industrial anomaly detection images]{\includegraphics[width=0.8\linewidth]{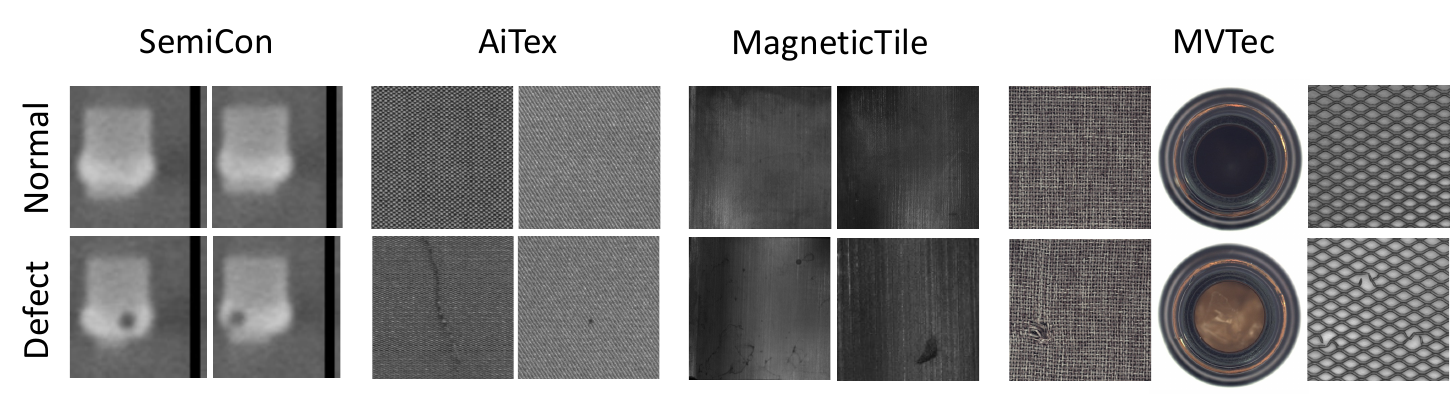}}
    \caption{(a) Selected examples from the 4 pairs~(8 species in total) in the Flowers17 dataset. Two images are shown for each species. (b) Examples of real-world datasets~(industrial images) with subtle variations between normal and anomalous samples.}
    \label{fig:examples}
\end{figure*}

\subsection{Datasets}

To demonstrate the effectiveness of our proposed anomaly detection method, we use a range of datasets belonging to one of two types, which we call controlled versus real-world. Their distinction will now be explained.

\subsubsection{Controlled Anomaly Detection Tasks}: In the controlled setup, we design anomaly detection tasks by altering existing classification benchmarks.
We first use the \textbf{Flowers17~\cite{nilsback2008automated}} dataset, which contains 17 classes of flower species, each with 80 images. This dataset helps to test the ability of our method to perform anomaly detection under a setting of great diversity within the set of normal data. Indeed, we see significant variations in data properties such as pose, color, shape and background within images belonging to the same class, i.e. high intra-class variance. Furthermore, flowers belonging to different classes can often exhibit strikingly similar properties, such as the daffodil and tulip classes. These two factors make for a particularly challenging anomaly detection task, where normal data can be highly diverse and anomalies can be subtly distinct from the normal class. 
To this end, we manually choose 4 pairs of two flower classes that are visually similar, shown in Fig.~\ref{fig:examples}~(a), which do not appear in ImageNet. One of the two classes is treated as the normal class, while the other is the anomaly, and we report results with each class set as the normal or anomalous class separately. {70 images from the normal class and 10 images from the abnormal class are reserved for the {test set} and we evaluate at 5-shot normal training samples.
This experiment has implications for the application of anomaly detection to agricultural purposes, e.g. detecting unwanted weeds that are visually similar to the wanted crops.
We also adapt the \textbf{CIFAR10/100-C~\cite{hendrycks2019benchmarking}} datasets to an anomaly detection problem. Similar to Flowers17, their classes exhibit high intra-class variation (within each normal class) and low inter-class variation (between normal and anomaly classes). We propose an anomaly detection protocol by treating all corrupted test samples as anomalies for each class. The dataset contains 5 levels of 15 different corruptions, e.g., various types of noise, blur and fog. For easy comparison we choose 9 types of level 4 corruptions, also shown in Fig.~\ref{fig:cifar}. 5000 clean testing samples from each class of original CIFAR10/100 dataset are used as normal testing samples, the corresponding corrupted samples of that class are treated as anomalous test samples. We evaluate at 10-shot normal training samples for both datasets.

\begin{figure}[!htb]
    \centering
    \includegraphics[width=1.02\linewidth]{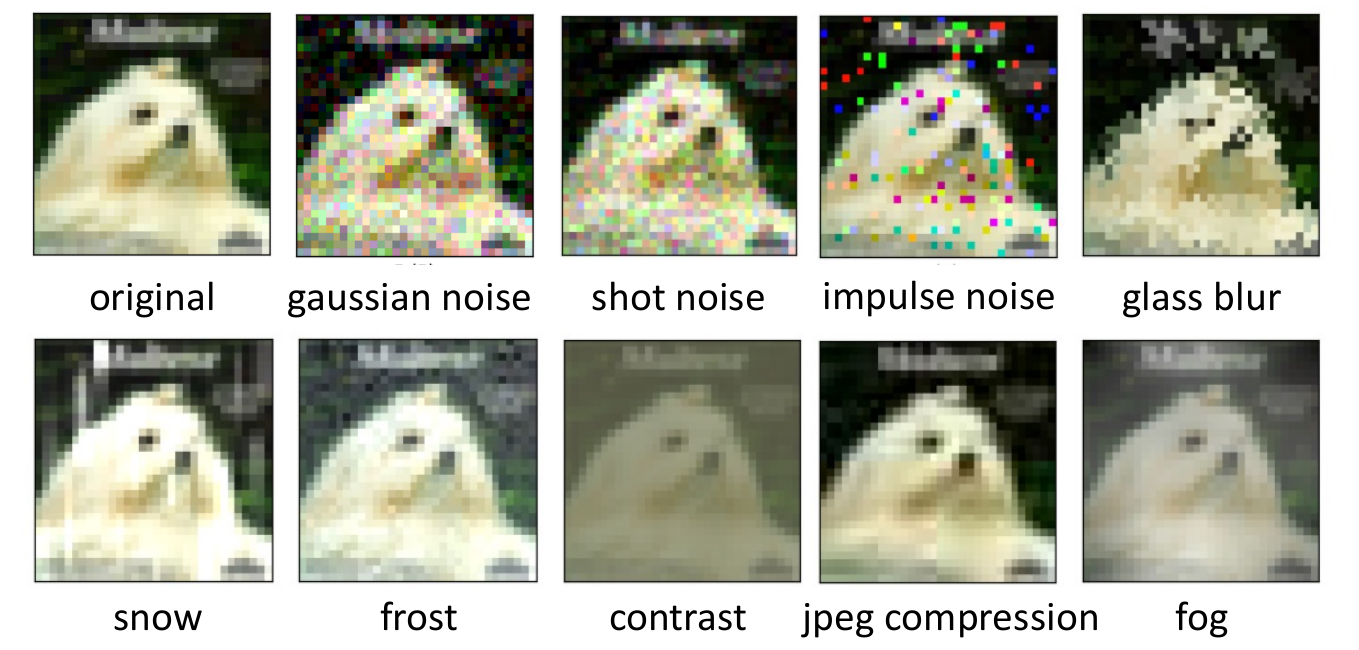}
    \caption{A selected example of a clean sample with different corruptions in the CIFAR10-C dataset.}
    \label{fig:cifar}
\end{figure}

\subsubsection{Real-world Anomaly Detection Tasks}: To evaluate our methodology on anomaly detection tasks which occur in more natural settings, we also use real-world datasets in our experiments.
Under this category, we evaluate 4 industrial anomaly detection datasets, including {MVTec Dataset~\cite{bergmann2019mvtec}}, {AITEX Dataset~\cite{silvestre2019public}}, {Magnetic Tile Defects Dataset~\cite{huang2020surface}} and {SemiCon Dataset~\cite{pahwa2021automated}}. Among these datasets foreground objects are often very well aligned with {highly} similar visual patterns, i.e. low intra-class variation and anomalies are caused by subtle defects, i.e. low inter-class variation. \textbf{MVTec Dataset~\cite{bergmann2019mvtec}} contains 15 object categories, including 10 non-texture object categories and 5 texture object categories. 
Each category contains 60-300 normal samples for training and 30-400 normal and defect samples for testing. We follow the few-shot settings from \cite{sheynin2021hierarchical} to create 2/5/10-shot anomaly detection protocols. 
 \textbf{AITEX Dataset~\cite{silvestre2019public}} is dedicated to detecting defects in textile fabric. This dataset consists of 140 normal sample images and 105 defect sample images with corresponding defect mask for localization. The original image resolution is 4096×256 pixels and the defects occupy a very small percentage of pixels. To allow for easier defect identification and mimicking a realistic defect identification procedure, we randomly crop out 5 patches of size $256\times256$ pixels from each image. Then 5/10/50-shot normal patches are randomly sampled as training examples. In total, test set consists of 600 normal patches and 105 defect patches.

 \textbf{Magnetic Tile Defects Dataset~\cite{huang2020surface}} consists of 6 different types of defect magnetic tile surface images, namely ``Blowhole'', ``Crack'', ``Fray'', ``Break'', ``Uneven'', and ``Free'' (no defects). There are 1344 images in total, of which 952 images contain defects. We randomly select 5/10/50-shot normal examples from the ``Free'' class as training data and evaluate on the remaining images for anomaly detection. \textbf{SemiCon Dataset~\cite{pahwa2021automated}} features 3D images collected from 3D X-ray microscopy (XRM) scans on integrated circuit packaging interconnects. The original dataset contains 53 3D images of memory dies and the anomaly detection task is to identify voids in the solder region: a large void to solder ratio may indicate defective products. To create an anomaly detection benchmark, we slice the 3D images to obtain 2D images. 5/10/50-shot normal images are randomly selected for training, 110 defect samples and 500 normal samples are used for testing.
 A few examples of anomalies from the industrial datasets are presented in Fig.~\ref{fig:examples}~(b)


\subsection{Competing Methods}

We compare against multiple anomaly detection methods, covering methods based on reconstruction~(AE and VAE), one-class classification~(DeepSVDD and CutPaste), generative models~(CFLOW-AD, TDG and RotNet) and self-supervised learning~(RotNet and CSI), in the experiments. We first benchmark auto-encoder (\textbf{AE}) employed in \cite{AE_SSIM}, which is trained to reconstruct input images and the difference between input and reconstructs measures the anomaly score. 
\textbf{VAE}~\cite{kingma2013auto} imposes constraints on the latent variables to be Gaussian. Multiple samples are drawn in the latent space and decoded to image space for measuring anomaly score~\cite{an2015variational}.
\textbf{DeepSVDD}~\cite{ruff2018deep} trains the network by forcing normal training samples to embed close to a cluster center. The learned cluster center can later serve as the prototype for anomaly detection by measuring the distance as anomaly score. 
\textbf{CutPaste}~\cite{li2021cutpaste} introduced CutPaste as an augmentation approach to synthesize negative examples for pretraining feature representation network. It is worth noticing that CutPaste mimics the real anomalies that would appear in the MVTec dataset, therefore the effectiveness of CutPaste does not generalize to other types of anomalies, e.g. differentiating flower species and corruptions as anomaly. 
 \textbf{CFLOW-AD~\cite{gudovskiy2022cflow}} adopted a conditional normalizing flow model for fast anomaly localization. We adapt CFLOW-AD to training on few-shot anomaly detection task. \textbf{TDG}~\cite{sheynin2021hierarchical} proposed to employ generative model for few-shot anomaly detection by differentiating image patches into either fake or one from a list of predefined augmentations.  \textbf{DifferNet}~\cite{rudolph2021same} estimates density through normalizing flow with a few supporting training samples. The above two approaches are compared on the MVTec dataset. 
 \textbf{RotNet}~\cite{golan2018deep} proposed a self-supervised pretraining approach by predicting the augmentations (rotations) applied. This approach is most effective in anomaly detection on natural semantic images and the advantage may {diminish} on industrial images where the pose of background is naturally more diverse. \textbf{CSI}~\cite{tack2020csi} proposed to contrast distribution shifted augmented images with original images to increase the gap between normal and abnormal samples. This is similar to maintaining only the negative pair loss proposed in this work.
 \textbf{COFT-AD~(w/o np)} optimizes on the combination of contrastive loss and cross-instance positive pair loss. Assuming prior knowledge of anomalies is available, \textbf{COFT-AD~(w/ np)} incorporates the negative pair loss and optimizes on the combination of all three losses.  Among these methods, CutPaste and COFT-AD~(w/ np) are built upon the prior knowledge of anomalies while other methods do not make explicit assumptions on how the anomalies appear.

\subsection{Experiment Details}

\noindent\textbf{Training Details}: For all experiments, we use the ResNet18~\cite{he2016deep} backbone for feature extraction. For all competing methods, we initialize backbone weights with ImageNet pretrained weights. We set the weight for cross-instance positive pair loss as $\lambda_{PP}=0.8$ and the weight for negative pair loss as $\lambda_{NP}=0.6$. We use the Adam optimizer~\cite{kingma2014adam} for all experiments with learning rate initialized to $3 \times 10^{-4}$, $\beta_1=0.9$ and $\beta_2=0.99$. We fix the batch size to 64, thus creating 64 pairs for contrastive training. To generate cross-instance pairs, we randomly permute the 64 images and each of the 64 image is paired with a randomly permuted one, resulting in 64 positive pairs. Similarly, we pair each image to its negatively augmented one to create another 64 negative pairs. For density model fitting, we use $N_A=10$. The area under the ROC (AUROC) is used to assess performance, where anomalies are treated as the positive class. 

\noindent\textbf{Data Augmentation}:
We aim to synthesize common variation within normal samples with positive augmentation and synthesize potential anomalies with negative augmentation.
For Flowers17 Dataset, we adopt  CutPaste~\cite{li2021cutpaste} for negative augmentation $t_n(\matr{X})$ and randomly affine transformation and color jittering for positive augmentation $t(\matr{X})$. For CIFAR10/100-C, positive augmentation includes random affine transformation while negative augmentation consists of blurring and randomly perturbing image brightness and contrast.
For all industrial dataset, positive augmentation consists of random affine transformation, blurring and grayscaling and negative augmentation consists of CutPaste~\cite{li2021cutpaste}.

\subsection{Results on Controlled Anomaly Detection Datasets}

In this section, we report on the evaluations for anomaly detection on controlled anomaly detection datasets. On the \textbf{Flowers17} dataset, we adapt ImageNet pretrained weights to each pair of flower species for anomaly detection. We report the results on the 4 pairs by swapping the two species as normal and abnormal in Tab.~\ref{tab:flowers}. We make the following observations from the results. First, we notice that COFT-AD outperforms competing methods on most pairs for anomaly detection. Importantly, without any prior knowledge on the anomalies, COFT-AD wins on 5 out of 8 pairs, suggesting the effectiveness of contrastive fine-tuning. Additionally, when negative pairs are included, COFT-AD~(w/ np) could be further improved, and there is a significant gap between CutPaste and COFT-AD where negative samples are synthesized in the same way. We argue that the advantage of COFT-AD is due to explicitly encouraging a tight cluster among normal samples. 
In contrast, CutPaste formulates anomaly detection as a binary classification task and could suffer from inaccurate synthesis of anomalies through cut and paste augmentation.

\begin{table*}[htbp]
  \centering
  \caption{5-shot anomaly detection performance on the Flowers17 dataset with Normal/Anomaly class setup. A~/~B with Norm.~/~Abn. indicates treating species A as normal and B as abnormal and vice versa for with Abn.~/~Norm. All numbers are reported as AUROC in $\%$.}
          \setlength\tabcolsep{2pt} 
          \resizebox{0.89\linewidth}{!}{
    \begin{tabular}{lcccccccc}
    \toprule
          & \multicolumn{2}{c}{\textbf{Crocus~/~Bluebell}} & \multicolumn{2}{c}{\textbf{Daffodil~/~Tulip}} & \multicolumn{2}{c}{\textbf{Lily Valley~/~Windflower}} & \multicolumn{2}{c}{\textbf{Buttercup~/~Cowslip}} \\
          & \multicolumn{1}{l}{Norm.~/~Abn.} & \multicolumn{1}{l}{Abn.~/~Norm.} & \multicolumn{1}{l}{Norm.~/~Abn.} & \multicolumn{1}{l}{Abn.~/~Norm.} & \multicolumn{1}{l}{Norm.~/~Abn.} & \multicolumn{1}{l}{Abn.~/~Norm.} & \multicolumn{1}{l}{Norm.~/~Abn.} & \multicolumn{1}{l}{Abn.~/~Norm.} \\
    \cmidrule(lr){1-1}\cmidrule(lr){2-3}\cmidrule(lr){4-5}\cmidrule(lr){6-7}\cmidrule(lr){8-9}
    \textbf{AE}    & 70.93 & 53.49 & 66.67 & 68.13 & 68.27 & 52.53 & 54.56 & 62.27 \\
    \textbf{VAE}   & 82.00 & 63.87 & 62.53 & 70.67 & 63.87 & 53.87 & 66.53 & 53.33 \\
    \textbf{DeepSVDD} & 77.47 & 48.00 & 49.02 & 45.06 & 66.93 & 71.07 & 61.47 & 60.93 \\
    \textbf{CSI}   & 74.13 & 46.05 & 55.87 & 53.87 & 72.80 & \textbf{81.60} & 64.53 & 88.67 \\
    \textbf{CFLOW-AD} & 83.73 & 52.93 & 67.60 & 44.27 & \textbf{81.60} & 72.93 & 76.40 & \textbf{90.07} \\
    \textbf{COFT-AD (w/o np)} & \textbf{89.47} & \textbf{64.80} & \textbf{68.00} & \textbf{71.49} & 80.27 & 76.39 & \textbf{83.33} & 89.47 \\
    \midrule
    \textbf{CutPaste} & 67.98 & 53.33 & 59.73 & 64.08 & 54.80 & 75.47 & 73.53 & 48.67 \\
    \textbf{COFT-AD (w/ np)} & \textbf{92.13} & \textbf{59.47} & \textbf{74.13} & \textbf{75.47} & \textbf{87.33} & \textbf{79.20} & \textbf{74.53} & \textbf{98.13} \\
    \bottomrule
    \end{tabular}%
    }
  \label{tab:flowers}%
\end{table*}%

\begin{table*}[htbp]
  \centering
  \caption{{10-shot anomaly detection on CIFAR10-C dataset.} All numbers are reported as AUROC in $\%$.}
        \setlength\tabcolsep{2pt} 
          \resizebox{0.85\linewidth}{!}{
    \begin{tabular}{p{10.0em}lllllllllll}
    \toprule 
    \multicolumn{1}{r}{} & \multicolumn{1}{p{3.5em}}{\textbf{Airpl.}} & \multicolumn{1}{p{3.5em}}{\textbf{Auto.}} & \multicolumn{1}{p{3.5em}}{\textbf{Bird}} & \multicolumn{1}{p{3.5em}}{\textbf{Cat}} & \multicolumn{1}{p{3.5em}}{\textbf{Deer}} & \multicolumn{1}{p{3.5em}}{\textbf{Dog}} & \multicolumn{1}{p{3.5em}}{\textbf{Frog}} & \multicolumn{1}{p{3.5em}}{\textbf{Horse}} & \multicolumn{1}{p{3.5em}}{\textbf{Ship}} & \multicolumn{1}
    {p{3.5em}}{\textbf{Truck}} & \multicolumn{1}{p{3.5em}}{\textbf{Avg.}} \\
    \midrule
    \textbf{AE} & 54.38 & 53.67 & 54.86 & 53.79 & 56.15 & 54.30 & 51.70 & 54.79 & 55.89 & 50.38 & 53.99 \\
    \textbf{VAE} & 57.47 & 52.17 & 56.32 & 56.11 & 58.67 & 57.09 & 59.18 & 54.06 & 56.90 & 50.61 & 55.86 \\
    \textbf{DeepSVDD}  & 63.27 & 61.64 & 60.75 & 60.71 & 61.54 & 62.11 & 63.00 & 59.83 & \textbf{63.72} & 67.29 & 62.37 \\
    \textbf{CSI} & 56.18 & 59.07 & 57.72 & 56.65 & 55.12 & 58.49 & 60.39 & 61.69 & 55.91 & 62.96 & 58.42\\
    \textbf{RotNet} & 69.21 & 53.21 & 66.03 & 64.00 & 67.30 & 67.84 & 70.20 & 70.84 & 59.05 & 63.04 & 64.07 \\
    \textbf{COFT-AD~(w/o np)} & \textbf{71.68} & \textbf{64.68} & \textbf{75.74} & \textbf{74.38} & \textbf{76.55} & \textbf{74.69} & \textbf{81.40} & \textbf{79.10} & 60.02 & \textbf{74.53} & \textbf{72.27} \\
    \midrule
    \textbf{CutPaste} & 58.62 & 73.01 & 70.56 & 75.57 & 68.65 & 71.32 & 72.26 & 70.82 & \textbf{72.53} & 66.79 & 70.01 \\
    \textbf{COFT-AD~(w/ np)} & \textbf{72.23} & \textbf{67.66} & \textbf{78.41} & \textbf{77.45} & \textbf{75.33} & \textbf{75.20} & \textbf{75.35} & \textbf{75.88} & 72.20 & \textbf{75.85} & \textbf{74.56} \\
    \bottomrule
    \end{tabular}%
    }
  \label{tab:CIFAR10}%
\end{table*}%

\begin{table*}[htbp]
  \centering
  \caption{Anomaly detection on CIFAR100-C dataset. Per super-class performance is reported. All numbers are reported as AUROC in $\%$.}
          \setlength\tabcolsep{1pt} 
          \resizebox{0.85\linewidth}{!}{
    \begin{tabular}{cccccccccccc}

    \toprule
          & \multicolumn{1}{c}{Sup.Cls. 0} & \multicolumn{1}{c}{Sup.Cls. 1} & \multicolumn{1}{c}{Sup.Cls. 2} & \multicolumn{1}{c}{Sup.Cls. 3} & \multicolumn{1}{c}{Sup.Cls. 4} & \multicolumn{1}{c}{Sup.Cls. 5} & \multicolumn{1}{c}{Sup.Cls. 6} & \multicolumn{1}{c}{Sup.Cls. 7} & \multicolumn{1}{c}{Sup.Cls. 8} & \multicolumn{1}{c}{Sup.Cls. 9} \\
    \midrule

    \textbf{AE} & 50.47 & 51.93 & 58.99 & 50.19 & 50.13 & 55.31 & 59.30 & 55.47 & 52.53 & 51.71 \\

    \textbf{VAE} & 50.43 & 55.85 & 57.55 & 56.59 & 58.7 & 55.30 & 56.30 & 56.92 & 57.62 & 56.69 \\
    
    \textbf{Deep SVDD} & 75.10 & 73.48 & 70.35 & 63.52 & 73.53 & 76.19 & 69.91 & 74.76 & \textbf{79.70} & 67.13 \\
    
    \textbf{CutPaste} & 75.96 & \textbf{69.59} & 70.51 & 75.71 & 72.50 & 74.14 & \textbf{76.22} & 76.69 & 71.79 & 78.31 \\
    \textbf{COFT-AD~(w/ np)} & \textbf{77.61} & 66.69 & \textbf{77.84} & \textbf{76.21} & \textbf{79.93} & \textbf{78.17} & 73.33 & \textbf{78.80} & 74.63 & \textbf{81.17} \\
    \midrule
    \multicolumn{1}{c}{Sup.Cls. 10} & \multicolumn{1}{c}{Sup.Cls. 11} & \multicolumn{1}{c}{Sup.Cls. 12} & \multicolumn{1}{c}{Sup.Cls. 13} & \multicolumn{1}{c}{Sup.Cls. 14} & \multicolumn{1}{c}{Sup.Cls. 15} & \multicolumn{1}{c}{Sup.Cls. 16} & \multicolumn{1}{c}{Sup.Cls. 17} & \multicolumn{1}{c}{Sup.Cls. 18} & \multicolumn{1}{c}{Sup.Cls. 19} & \multicolumn{1}{c}{Avg} \\
    \midrule

    52.57 & 55.15 & 51.61 & 57.44 & 59.12 & 50.91 & 55.66 & 56.36 & 51.96 & 55.18 & 54.10\\
    60.99 & 50.21 & 51.79 & 52.16 & 60.96 & 57.27 & 61.95 & 56.61 & 57.42 & 59.26 & 56.53\\
    70.00 & 73.17 & 60.82 & 56.91 & 66.36 & 65.81 & 60.60 & 69.88 & 70.52 & 71.30 & 69.24 \\
    80.28 & \textbf{75.83} & 77.92 & 78.56 & 70.87 & 78.84 & 72.83 & 71.63 & 72.97 & 75.61 & 74.84  \\

    \textbf{81.27} & 75.59 & \textbf{80.11} & \textbf{79.98} & \textbf{75.56} & \textbf{80.04} & \textbf{73.65} & \textbf{77.54} & \textbf{77.16} & \textbf{78.13} & \textbf{77.17} \\

    \bottomrule
    \end{tabular}%
    }
  \label{tab:addlabel}%
\end{table*}%

We further report benchmarks on \textbf{CIFAR10-C} and \textbf{CIFAR100-C} with 10-shot training samples. In a similar fashion to the Flowers17 dataset, we adapt ImageNet pretrained weights to each individual semantic class. For CIFAR100-C, we choose the 20 super-classes as the semantic class for simplicity. We present the anomaly detection results on each semantic category of CIFAR10-C in Tab.\ref{tab:CIFAR10}. We first observe from the results that, on average, our method~(COFT-AD), both with and without prior knowledge of negative pairs~(w/ np or w/o np), outperforms competing methods by a clear margin. The closest competing method, CutPaste, is $4\%$ lower than COFT-AD~(w/ np). This is in contrast to the extraordinary performance demonstrated on MVTec by CutPaste, as shown in the following section. We attribute this to the fact that the corruptions in CIFAR10-C are diverse and may not be easily synthesized by the augmentation methods, i.e. CutPaste, specifically tailored for the MVTec dataset. 
We further benchmark on CIFAR100-C and compare with DeepSVDD and CutPaste. We draw similar conclusions from the results. Our method is still stronger than CutPaste with a clear margin. This is caused by the mismatch between the negative samples synthesized by CutPaste and corruptions~(anomalies) in the dataset.

\begin{table*}[!htbp]
  \centering
  \caption{Few-shot anomaly detection on MVTec dataset. Per-category AUROC is reported for all competing methods. All numbers are in $\%$. The results of DiffNet$^*$ and TDG$^*$ are derived from \cite{sheynin2021hierarchical}, where $-$ indicates per-class results are not available.}
      \setlength\tabcolsep{4pt} 
        \resizebox{1.0\linewidth}{!}{    \begin{tabular}{clllllllllllllllll}
    \toprule
          &       & \textbf{bottle} & \textbf{cable} & \textbf{caps.} & \textbf{hazel.} & \textbf{metal.} & \textbf{pill} & \textbf{screw} & \textbf{tooth.} & \textbf{transis.} & \textbf{zipper} & \textbf{carpet} & \textbf{grid} & \textbf{leather} & \textbf{tile} & \textbf{wood} & \textbf{avg.} \\
    \midrule
    \multirow{5}[2]{*}{\begin{sideways}\textbf{2 shot}\end{sideways}} & \textbf{AE} & 73.49 & 64.22 & \textbf{62.43} & 73.54 & 35.97 & 75.19 & 35.56 & 73.33 & 48.92 & 40.73 & 22.11 & 45.13 & 31.52 & 73.35 & 58.42 & 53.26 \\
          & \textbf{VAE} & 68.73 & 61.83 & 60.47 & \textbf{73.82} & 41.20 & 76.08 & 39.98 & 72.22 & 68.83 & 38.16 & 25.08 & 40.85 & 37.40 & 72.69 & 44.91 & 54.75 \\
          & \textbf{DeepSVDD} & 85.79 & \textbf{66.06} & 51.62 & 53.39 & 50.44 & \textbf{77.27} & 51.23 & 69.72 & 58.04 & 59.30 & 70.06 & 38.10 & 40.42 & \textbf{81.02} & 51.23 & 60.28 \\
          & \textbf{COFT-AD~(w/o np)} & \textbf{88.06} & 59.05 & 58.82 & 60.77 & \textbf{68.51} & 65.55 & \textbf{56.48} & \textbf{73.63} & \textbf{72.84} & \textbf{70.74} & \textbf{76.00} & \textbf{61.18} & \textbf{63.11} & 68.25 & \textbf{77.85} & \textbf{68.06} \\
          \cmidrule{2-18}
          & \textbf{CutPaste} & 86.39 & 64.59 & \textbf{61.56} & \textbf{73.59} & 49.92 & \textbf{66.91} & 41.93 & \textbf{82.04} & 55.53 & 59.97 & 52.26 & 46.25 & \textbf{83.82} & \textbf{71.28} & \textbf{84.79} & 65.38 \\
          & \textbf{COFT-AD~(w/ np)} & \textbf{90.95} & \textbf{66.55} & 59.38 & 68.82 & \textbf{68.96} & 66.68 & \textbf{54.90} & 74.81 & \textbf{74.33} & \textbf{72.82} & \textbf{73.18} & \textbf{61.77} & 65.28 & 69.64 & 79.85 & \textbf{69.86} \\
    \midrule
    \multirow{7}[2]{*}{\begin{sideways}\textbf{5 shot}\end{sideways}} & \textbf{AE} & 76.59 & 65.59 & 72.92 & 73.64 & 49.61 & 76.73 & 40.32 & 75.00 & 64.79 & 59.95 & 38.76 & 42.61 & 43.72 & 66.56 & 74.04 & 61.39  \\
          & \textbf{VAE} &  70.24 & 62.52 & 73.08 & 74.57 & 41.96 & 76.55 & 41.55 & 79.51 & 72.92 & 59.50 & 39.32 & 56.16 & 45.99 & 60.22 & 90.16 & 63.05 \\
          & \textbf{DeepSVDD} & 86.19 & 68.29 & 60.23 & 54.07 & 52.16 & 78.42 & 52.15 & 81.39 & 69.04 & 74.80 & 51.54 & 51.55 & 58.20 & {82.83} & \textbf{93.51} & 67.59 \\
          & \textbf{CSI} & 80.55 & 60.48 & 62.07 & 74.40 & 59.83 & 69.29 & 32.71 & 79.44 & 55.71 & 63.71 & 56.17 & 39.53 & 51.56 & 55.38 & 75.96 & 61.12\\
          & \textbf{CFLOW-AD} & \textbf{98.17} & \textbf{81.65} & 73.12 & 88.25 & 74.88 & 68.22 & 45.09 & 82.50 & \textbf{84.79} & \textbf{83.53} & 73.48 & 50.96 & \textbf{87.84} & \textbf{91.59} & 92.37 & \textbf{78.43} \\
          & \textbf{DifferNet}$^*$ & -     & -     & -     & -     & -     & -     & -     & -     & -     & -     & -     & -     & -     & -     & -     & 72.10 \\
          & \textbf{TDG}$^*$ & -     & -     & -     & -     & -     & -     & -     & -     & -     & -     & -     & -     & -     & -     & -     & 77.90 \\
          & \textbf{COFT-AD~(w/o np)} & {94.58} & {70.67} & \textbf{74.06} & {78.29} & \textbf{79.66} & \textbf{81.56} & \textbf{62.70} & \textbf{87.18} & {78.08} & {75.03} & \textbf{84.67} & \textbf{62.31} & {74.90} & 78.34 & 89.66 & {78.11}\\
          \cmidrule{2-18}
          & \textbf{CutPaste} & \textbf{98.41} & \textbf{80.32} & 69.20 & \textbf{89.90} & 72.24 & \textbf{82.85} & 59.13 & \textbf{90.89} & 68.56 & 68.89 & 73.13 & 49.94 & \textbf{83.93} & \textbf{91.50} & \textbf{96.08} & 78.33 \\
          & \textbf{COFT-AD~(w/ np)} & 95.73 & 78.05 & \textbf{74.26} & 86.89 & \textbf{78.57} & 81.31 & \textbf{62.26} & 88.81 & \textbf{74.80} & \textbf{75.70} & \textbf{79.90} & \textbf{61.85} & 77.92 & 75.34 & 90.12 & \textbf{78.76} \\
    \midrule
    \multirow{7}[1]{*}{\begin{sideways}\textbf{10 shot}\end{sideways}} & \textbf{AE} & 81.91 & 69.34 & 73.54 & 74.14 & 57.72 & 78.40  & 50.07 & 93.11 & 66.49 & 60.29 & 41.43 & 49.96 & 45.07 & 72.62 & \textbf{95.70} & 64.92 \\
          & \textbf{VAE} & 82.06 & 64.05 & 73.59 & 75.04 & 56.74 & 78.07 & 50.03 & 91.66 & 73.41 & 60.19 & 44.78 & 56.05 & 47.45 & 76.55 & 94.56 & 67.02 \\
          & \textbf{DeepSVDD} & 86.75 & 68.85 & 65.05 & 74.04 & 70.82 & 78.42 & 53.13 & 86.39 & 69.08 & 77.28 & 52.07 & 51.71 & 58.27 & {86.33} & 94.02 & 71.48 \\
          & \textbf{CSI} & 83.44 & 62.65 & 64.30 & 77.07 & 61.98 & 71.78 & 33.89 & 82.30 & 57.71 & 66.00 & 58.19 & 40.95 & 53.41 & 57.37 & 78.69 & 63.32\\
          & \textbf{CFLOW-AD} & \textbf{99.50} & \textbf{84.58} & 75.75 & \textbf{91.42} & 77.57 & 70.67 & 46.71 & 85.47 & \textbf{87.84} & \textbf{86.53} & 76.12 & 52.79 & \textbf{91.00} & \textbf{94.88} & {95.69} & \textbf{81.10} \\
          & \textbf{DifferNet}$^*$ & -     & -     & -     & -     & -     & -     & -     & -     & -     & -     & -     & -     & -     & -     & -     & 73.60 \\
          & \textbf{TDG}$^*$ & -     & -     & -     & -     & -     & -     & -     & -     & -     & -     & -     & -     & -     & -     & -     & 78.00 \\
          & \textbf{COFT-AD~(w/o np)} & {97.84} & {71.07} & \textbf{79.23} & {78.72} & \textbf{80.57} & \textbf{82.81} & \textbf{61.85} & \textbf{95.12} & {87.21} & {83.71} & \textbf{84.73} & \textbf{63.10} & {76.88} & 81.14 & 89.69 & {80.92}\\
          \cmidrule{2-18}
          & \textbf{CutPaste} & 98.71 & 81.83 & \textbf{83.15} & \textbf{94.47} & \textbf{88.92} & \textbf{85.63} & \textbf{64.55} & 91.50  & 70.01 & \textbf{86.90}  & \textbf{83.92} & 55.13 & \textbf{99.50}  & \textbf{91.81} & 96.21 & \textbf{84.82} \\
          & \textbf{COFT-AD~(w/ np)} & \textbf{99.03} & \textbf{83.49} & 78.38 & 87.25 & 79.47 & 82.55 & 63.92 & \textbf{94.71} & \textbf{87.21} & 84.45 & 79.95 & \textbf{63.14} & 78.88 & 81.14 & \textbf{97.15} & 82.71 \\
          \bottomrule
    \end{tabular}%
    }
  \label{tab:MVTec}%
\end{table*}%

\begin{table*}[!htb]
  \centering
  \caption{Few-shot defect identification results on additional three industry image datasets. AUROC is reported as evaluation metrics. All numbers are in ($\%$).}
      \setlength\tabcolsep{5pt} 
        \resizebox{0.7\linewidth}{!}{
    \begin{tabular}{lrrr|rrr|rrr}
    \toprule
          & \multicolumn{3}{c}{\textbf{SemiCon}} & \multicolumn{3}{c}{\textbf{AITEX}} & \multicolumn{3}{c}{\textbf{Magnetic Tile}} \\
          & \multicolumn{1}{l}{\textbf{5-shot}} & \multicolumn{1}{l}{\textbf{10-shot}} & \multicolumn{1}{l}{\textbf{50-shot}} & \multicolumn{1}{l}{\textbf{5-shot}} & \multicolumn{1}{l}{\textbf{10-shot}} & \multicolumn{1}{l}{\textbf{50-shot}} & \multicolumn{1}{l}{\textbf{5-shot}} & \multicolumn{1}{l}{\textbf{10-shot}} & \multicolumn{1}{l}{\textbf{50-shot}} \\
\cmidrule(lr){1-1}\cmidrule(lr){2-4}\cmidrule(lr){5-7}\cmidrule(lr){8-10}    \textbf{AE}    & 65.38 & 70.24 & 71.74 & 47.01 & 60.59 & 63.30 & 51.58 & 52.66 & 54.70 \\
    \textbf{VAE}   & 72.42 & 73.53 & 80.14 & 52.99 & 66.36 & 67.88 & 51.90 & 53.40 & 54.30 \\
    \textbf{DeepSVDD} & 52.02 & 71.40 & 79.09 & 70.15 & 74.40 & \textbf{80.07} & 54.80 & 55.90 & 57.73 \\
    \textbf{RotNet} & 55.13 & 75.55 & 80.45 & 71.20 & 75.19 & 80.02 & 55.69 & 57.14 & 60.00 \\
\textbf{CutPaste} & 48.71 & 71.60 & 75.03 & 50.23 & 69.58 & 79.83 & 57.82 & 61.00 & 62.14 \\
    \textbf{COFT-AD~(w/o np)}  & \textbf{78.87} & \textbf{80.72} & \textbf{82.00} & \textbf{73.44} & \textbf{77.45} & 78.14 & \textbf{58.23} & \textbf{61.93} & \textbf{63.55} \\
    \bottomrule
    \end{tabular}%
    }
  \label{tab:AddIndustry}%
\end{table*}%

\subsection{Results on Real-world Datasets}

In this section, we explore identifying defects on real-world industrial images. We first evaluate the few-shot anomaly detection performance on \textbf{MVTec} dataset with results presented in Tab.~\ref{tab:MVTec}.
We make the following observations. First, without any specific prior knowledge on the anomalies, our method (COFT-AD~(w/o np)) outperforms all competing methods by a clear margin. Furthermore, with prior knowledge on the potential anomalies, our method (COFT-AD~(w/ np)) still outperforms CutPaste with the same negative augmentations in the 2-shot and 5-shot settings. We are only slightly behind CutPaste in the 10-shot case. Both observations suggest the effectiveness of contrastive adaptation and cross-instance positive pair loss. We further observe that CutPaste exhibits a significant lead on leather, wood and toothbrush images. We attribute this to the fact that these categories contain many anomalies that can be synthesized from CutPaste and scar augmentation: the ``cut'' defect for the ``leather'' category, the ``scratch'' defect for ``wood'' and scar like defects for ``toothbrush''. In contrast, COFT-AD relies more on adaptation from the pretrained model and learning from few-shot normal samples. As a result, its performance is generally better on more diverse types of objects: in the 2-shot setting, COFT-AD outperforms CutPaste on 8 of the 15 categories while CutPaste is winning on 4/15 categories. Finally, we make the observation that our proposed method does not benefit substantially upon more available normal training samples. We believe this is due to the repetitive patterns of normal data causing low intra-class variation within normal data distribution. For example, the normal samples of MVTec dataset all look very similar, thus a few normal samples are enough for learning good model for predicting anomalies.

We further evaluate defect identification performance on another three industrial datasets, namely \textbf{SemiCon}, \textbf{AITEX} and \textbf{MagneticTile}, with results presented in Tab.~\ref{tab:AddIndustry}. The following observations are drawn from the results. First, without any prior knowledge, COFT-AD~(w/o np) achieves the state-of-the-art performance under lower budgets of available training samples (5 and 10 shots) on all three datasets. It is only slightly behind DeepSVDD at 50-shot on AITEX. These results suggest adapting pretrained models to target domain is effective for realistic industrial image defect identification tasks. Second, methods demonstrating strong performance on MVTec dataset may not generalize to other types of defects. For example, while CutPaste is one of the best performing methods on MVTec, its performance on SemiCon and AITEX is much worse than more traditional approaches. One potential reason for this poor performance is that the synthesized negative samples used in CutPaste are not representative of the defects in SemiCon and AITEX datasets. 
\begin{table*}[!htbp]\centering
\caption{Ablation study on CIFAR10-C 10-shot FSAD. CIPP stands for cross-instance positive pair.}\label{tab:ablation}
          \setlength\tabcolsep{3pt}
\resizebox{0.75\linewidth}{!}{
\begin{tabular}{ccccccc}\toprule
Pretrained Weights & Contrast. Train & Positive Pair Loss & Negative Pair Loss & L2 Norm & ResNet18 & ResNet50 \\
\cmidrule(lr){1-1}\cmidrule(lr){2-2}\cmidrule(lr){3-3}\cmidrule(lr){4-4}\cmidrule(lr){5-5}\cmidrule(lr){6-6}\cmidrule(lr){7-7}
- &- &- &- &\checkmark & 52.54 & 54.04\\
ImageNet &- &- &- & \checkmark & 67.32 & 66.37 \\
ImageNet & \checkmark &- &- &\checkmark & 70.05 & 68.14\\
ImageNet & \checkmark &F.C.~\cite{ruff2018deep} &- &\checkmark & 68.21 & 70.00\\
ImageNet & \checkmark &C.I.P.P. &- &\checkmark & 73.68 & 75.00\\
ImageNet & \checkmark &C.I.P.P. &\checkmark &\checkmark & \textbf{74.56} & \textbf{76.69}\\
ImageNet & \checkmark &C.I.P.P. &\checkmark &- & 72.11 & 74.18\\

\bottomrule
\end{tabular}
}
\end{table*}

\subsection{Ablation Study}

In this section, we investigate the effectiveness of the individual components using the CIFAR10-C dataset as the anomalies are well-controlled. In particular, we demonstrate the importance of contrastive training on few-shot target domain normal samples (Contrast. Train), incorporating cross-instance positive pair loss (Positive Pair Loss) and incorporating the negative pair loss (Negative Pair Loss). We further evaluate incorporating L2 normalization (L2 Norm) during anomaly detection density model fitting and inference. We present the ablation study results in Tab.~\ref{tab:ablation}. We make the following observations from the results. First, as we expected, reusing ImageNet pretrained weights for downstream anomaly detection yields significant improvement in performance ($52.54\%\rightarrow67.32\%$). This suggests the significance of a good representation for anomaly detection. 
Adapting pretrained model to the target distribution through contrastive training further improves $2\%$ in average ($67.32\%\rightarrow70.05\%$). 
To encourage feature embedding suitable for density-based anomaly detection, we further incorporate the positive pair loss and this again yields additional $3\%$ improvement ($70.05\%\rightarrow73.68\%$). As an alternative approach, one could encourage all normal samples' features to embed close to a fixed cluster center~(F.C.) following \cite{ruff2018deep}. However, as the cluster center must be fixed through the first forward pass, this could pose too much constraint on the representation learning and yield inferior results ($68.21\%$). Finally, when we combine negative pair loss, this gives a final boost of performance to $74.56\%$. We also hypothesize that L2 normalization on feature representation is necessary and the ablation study validate the hypothesis. By removing the L2 normalization on anomaly inference features, the performance drops from $74.56\%$ to $72.11\%$, indicating the normalization is essential to fitting better density model and distance-based anomaly detection. Additionally, we replaced the backbone model with ResNet50 in the same controlled experiments to validate that our method performs well across different backbones. It is noteworthy that using only ImageNet pretrained weights for ResNet50 yields lower performance compared to ResNet18. This is attributed to the fact that an overly expressive representation may not necessarily be advantageous for anomaly detection\cite{reiss2023free}.

\section{Further Analysis}

In this section, we provide additional evaluations on qualitative results and discuss when incorporating negative samples should be employed.

\subsection{Qualitative Analysis}

We present qualitative results on the {Flowers17} dataset to examine both successful and failure cases in Fig.~\ref{fig:qualitative}. Specifically, we treat ``tulip'' as normal class and ``daffodil'' as abnormal class. We first visualize the 5-shot training samples in the first row of Fig.~\ref{fig:qualitative}, which covers different stages of the flower~(bud to full-blown). For each testing sample, we calculate a normalized anomaly score as the percentile among all testing samples. In the first row of normal testing samples~(TEST~-~Tulip~(Normal)), we show the normal testing samples with low anomaly scores. We clearly observe that testing samples with similar pose and shapes to the few-shot training samples are predicted as normal with high confidence~(low anomaly score). In the second row of images, we further present normal testing samples with high anomaly score. We observe these false positive samples are either taken from different viewpoints~(2-4 cols) or look significantly different~(round petals in 6-7 cols) from the few-shot training samples. Lastly, we present the abnormal testing samples in the last row of Fig.~\ref{fig:qualitative}. The first example~(1st col) is predicted by most methods with low anomaly score because the petals are pointy, resembling the tulip flowers in the training set. Other examples with round petals~(3-7 cols) are well predicted by COFT-AD with high anomaly score.

\begin{figure*}
    \centering
    \includegraphics[width=0.99\linewidth]{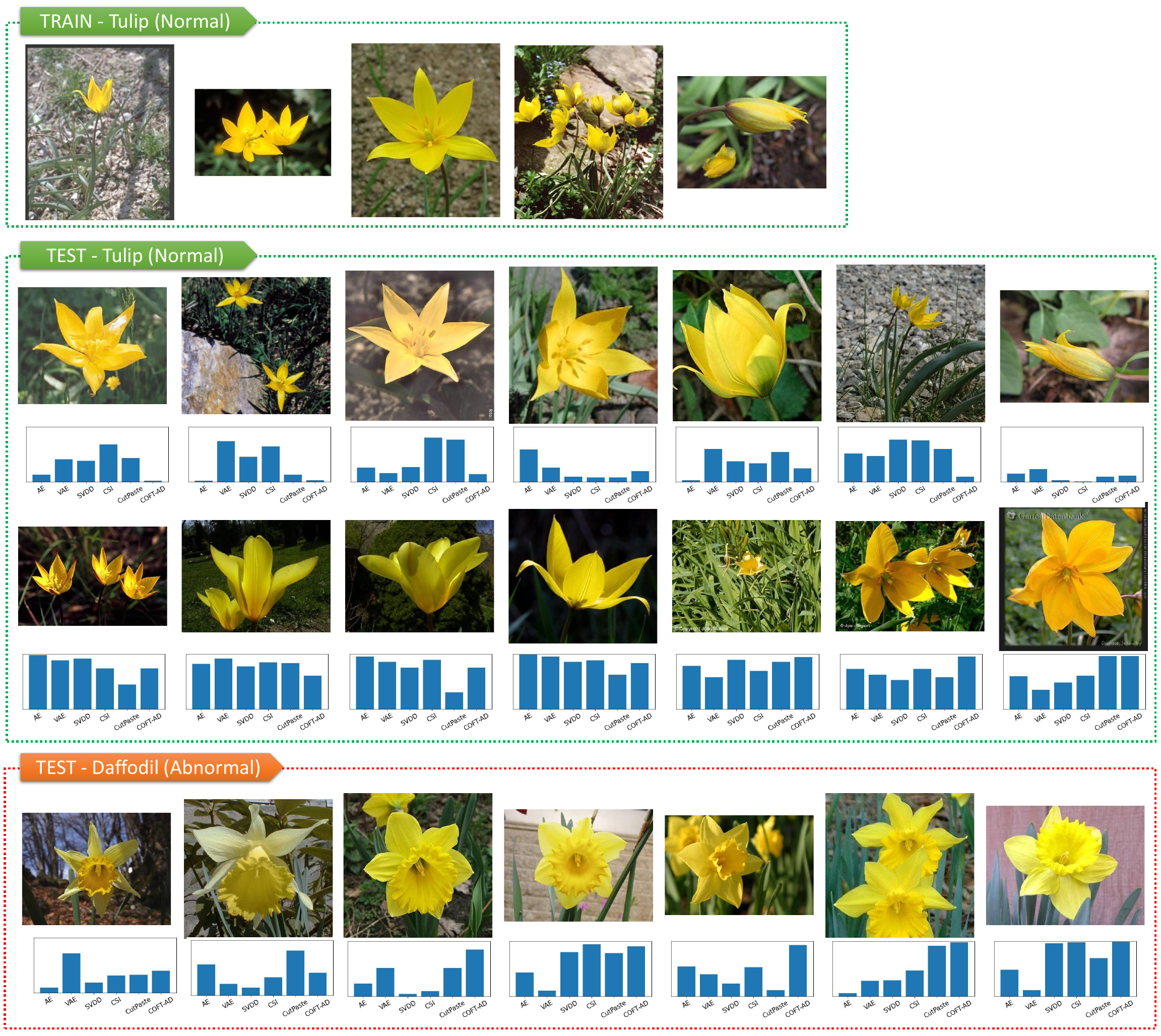}
    \caption{Qualitative evaluation of anomaly detection results on Flowers17 dataset. Each testing sample is accompanied with normalized anomaly scores predicted by different competing methods.}
    \label{fig:qualitative}
\end{figure*}

\subsection{Qualitative Examination of Representation Learning}
We provide qualitative observations into the advantage of incorporating the proposed losses through t-SNE visualization~\cite{van2008visualizing} of test data representations. Specifically, we randomly select 1,500 testing samples from 
CIFAR10-C dataset for visualization. The feature points are projected into 2D space and visualized in Fig.~\ref{fig:TSNE}. The feature embedding with ImageNet pretrained weights only, (a) w/o contrastive fine-tuning, shows a substantial overlap between normal and abnormal samples. When contrastive training is applied, (b) w/ contrastive fine-tuning, we observe a clear seperation between normal and abnormal samples. When additional positive pair loss, (c) w/ pp Loss, and negative pair loss, (d) w/ np Loss, are incorporated, the normal samples are further grouped into a tighter cluster with larger distinction between normal and abnormal samples.

\begin{figure}[!htb]
    \centering
    \subfloat[w/o contrastive fine-tuning]{\includegraphics[width=0.49\linewidth]{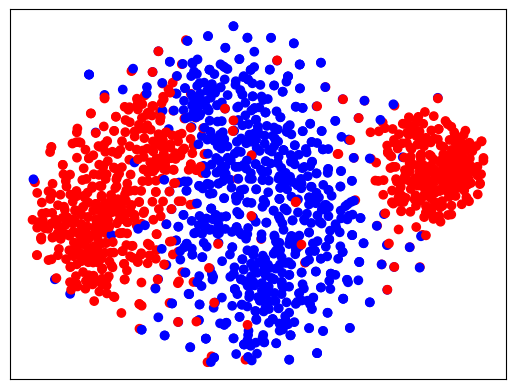}}
    \subfloat[w/ contrastive fine-tuning]{\includegraphics[width=0.49\linewidth]{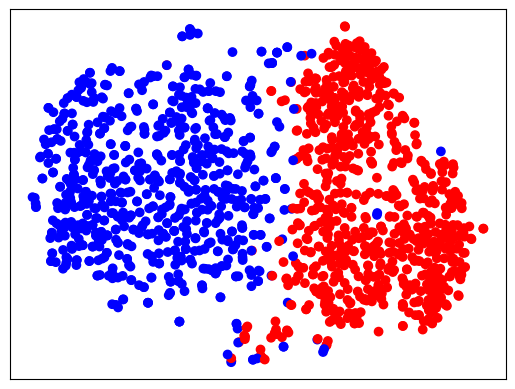}}\\
    \subfloat[w/ pp Loss]{\includegraphics[width=0.49\linewidth]{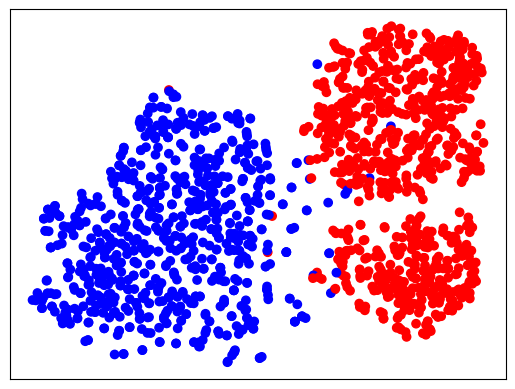}}
    \subfloat[w/ np Loss]{\includegraphics[width=0.49\linewidth]{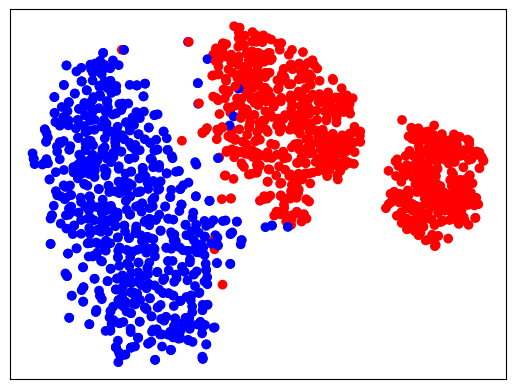}}
    \caption{T-SNE visualization of anomaly detection testing data on selected testing samples from CIFAR10-C dataset by incrementally incorporating different loss terms. Blue and red colors indicate normal and abnormal samples respectively.}
    \label{fig:TSNE}
\end{figure}

\subsection{Incorporating Negative Examples}
As we discussed, incorporating negative example during adaptation is not always beneficial. The advantage hinges on whether prior knowledge on abnormal sample distribution is available or not. To verify this hypothesis, we evaluate incorporating negative pair loss on 3 industry datasets, SemiCon, AITEX and Magnetic Tile. SemiCon dataset features anomalies that are quite different from MVTec, while AITEX and Magnetic Tile are relatively more similar to the texture categories of MVTec. We choose CutPaste~\cite{li2021cutpaste} as negative augmentation for these 3 datasets. The results in Tab.~\ref{tab:IncorpNP} demonstrate that when the negative augmentation is substantially different from the real anomalies, e.g. the void circles in SemiCon dataset, incorporating negative pair loss with inappropriate augmentation may harm performance ($78.87\rightarrow62.97\%$). On the contrary, when anomalies can be simulated, even imperfectly, e.g. Magnetic Tile dataset, incorporating negative pair loss will further improve performance. Overall, we conclude that incorporating negative pair loss is most helpful when prior knowledge on the potential anomalies is concrete and anomalies can be simulated through negative augmentation. Our model without negative pair loss is suitable for tasks without prior knowledge or when generating negative augmentation is difficult.

\begin{table}[!htp]\centering
\caption{Effect of incorporating negative pair loss with CutPaste augmentation on 10-shots industry image datasets.}\label{tab:IncorpNP}
\setlength\tabcolsep{3pt} 
\resizebox{0.85\linewidth}{!}{
\begin{tabular}{lccc}\toprule
 & SemiCon & AITEX & Magnetic Tile\\\midrule
\textbf{COFT-AD~(w/o np)} & 78.87 &  73.44 & 58.23\\
\textbf{COFT-AD~(w/ np)} & 76.29 &  65.54 & 59.45 \\
\bottomrule
\end{tabular}
}
\end{table}

\subsection{Comparison of Anomaly Scoring Function}
In this section, we further validate the effectiveness of contrastive learning-extracted features by comparing various anomaly scoring functions. We conducted comparisons with another classic non-parametric classifier, KNN~\cite{pang2018learning}, Deep Isolation Forest\cite{Xu_2023}, which is specifically designed for anomaly detection, and PatchCore\cite{roth2022towards}. The results\ref{tab:scoringFunction} demonstrate that the features obtained through contrastive learning exhibit stable performance across different anomaly detection models. Specifically for PatchCore, the method is designed for industrial defect detection and not applicable on dataset with high intra-class variations. When the intra-class variation is excessively large, the memory bank may not adequately represent the dataset's features, and anomalies can not be well-captured by local features.

\begin{table}[!htp]\centering
\caption{Comparison between different anomaly scoring functions. Experiments on Flower17 dataset, Daffodil (normal) / Tulip (abnormal). }\label{tab:scoringFunction}
\setlength\tabcolsep{3pt}
\resizebox{0.9\linewidth}{!}{
\begin{tabular}{lcccc}\toprule
 & Gaussian & kNN~\cite{pang2018learning} & Isolation Forest\cite{Xu_2023} & PatchCore\cite{roth2022towards}\\\midrule
ImageNet Pretrained & 52.40 & 44.27 & 50.80 & 53.20\\
\textbf{COFT-AD~(w/o np)} & 68.00 & 60.55  &  66.00 & 57.45\\
\textbf{COFT-AD~(w/ np)} & 74.13 & 77.47  & 73.33 & 59.90\\
\bottomrule
\end{tabular}
}
\end{table}

\subsection{Distribution of Anomaly Scores}
{
In this section, we compare different ablated models through visualizing the distribution of anomaly scores on CIFAR10-C few-shot anomaly detection task. Specifically, we compare a) w/o contrastive finetuning; b) w/ contrastive finetuning; c) method b), additionally w/ PP loss; and d) method c), additionally w/ NP loss. We make the following observations from the results in Fig.~\ref{fig:AnoScoreDist}. First, directly using the model pretrained on ImageNet, a) w/o contrastive fine-tuning, fails to differentiate normal and abnormal samples as the two distributions are almost the same. With contrastive fine-tuning on target data, b) w/ contrastive fine-tuning, we see a clear gap between normal and abnormal distributions, suggesting better anomaly detection performance. By further incorporating the positive pair loss, c) w/pp loss, the gap between two distributions becomes more significant. Finally, incorporating negative pair loss, d) w/ np loss, is most effective for differentiating normal from abnormal samples.}

\begin{figure}
    \centering
    \subfloat[w/o contrastive fine-tuning]{\includegraphics[width=0.5\linewidth]{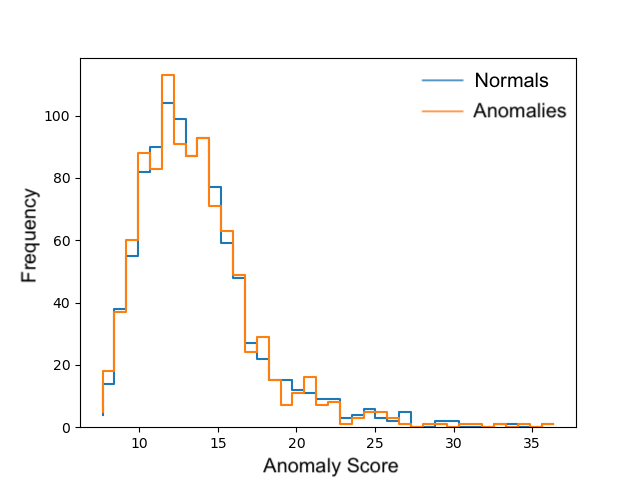}}
    \subfloat[w/ contrastive fine-tuning]{\includegraphics[width=0.5\linewidth]{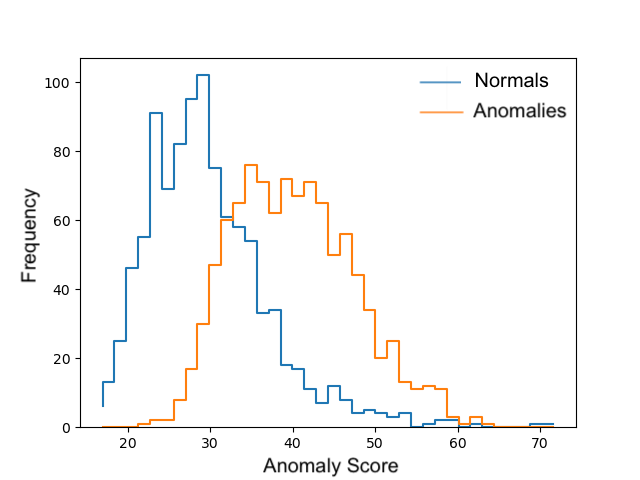}}\\
    \subfloat[w/ pp loss]{\includegraphics[width=0.5\linewidth]{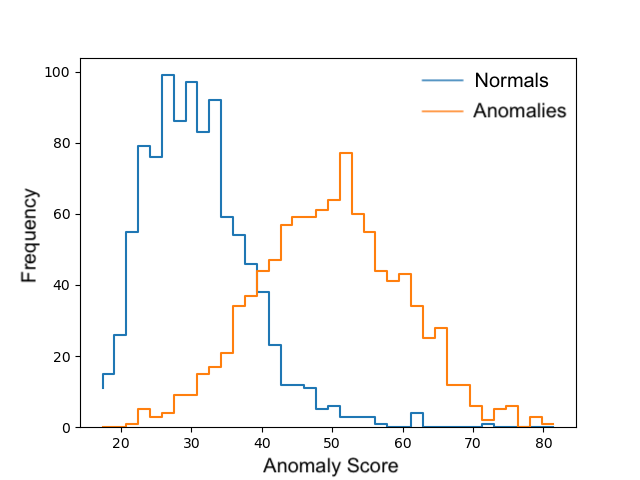}}
    \subfloat[w/ np loss]{\includegraphics[width=0.5\linewidth]{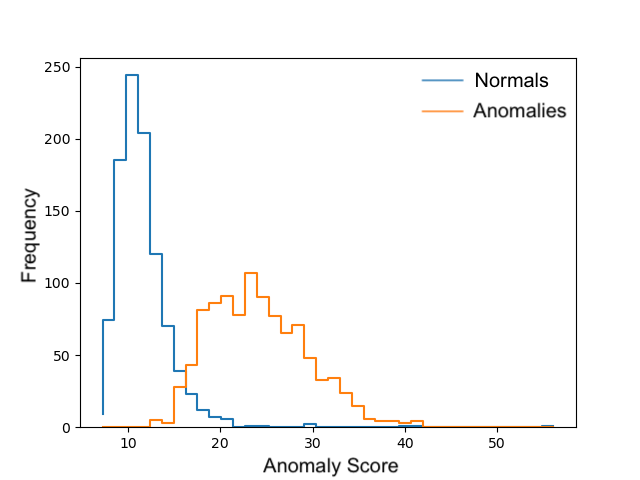}}
    \caption{Comparing different ablated models through anomaly score distributions.}
    \label{fig:AnoScoreDist}
\end{figure}

\subsection{Imbalanced Anomaly Detection}

{The ratio between normal and abnormal samples could have an impact of anomaly detection performance. Since the ratio between normal and abnormal samples does not affect model training, we implement a controlled experiment where we manually decrease the number of anomalies in the SemiCon dataset. Specifically, we fix the normal samples to 500 and randomly subsample 5, 10, 20 and 50 abnormal samples, respectively indicating $1\%$, $2\%$, $4\%$, $10\%$, abnormal to normal ratios, for evaluation. The results in Tab.~\ref{tab:imbalance} demonstrate the superiority of proposed method against existing competing methods. }

\begin{table}[!htp]\centering
\caption{Evaluation of anomaly detection under imbalanced anomaly samples.}\label{tab:imbalance}
\scriptsize
\begin{tabular}{lrrrrr}\toprule
\#Anomalies &5~(1\%) &10~(2\%) &20~(4\%) &50~(10\%) \\\midrule
\textbf{DeepSVDD} &52.01 &49.71 &49.97 &52.40 \\
\textbf{CutPaste} &86.31 &83.82 &72.82 &74.31 \\
\textbf{VAE} &80.56 &78.12 &67.67 &73.91 \\
\textbf{AE} &62.12 &65.52 &76.83 &74.39 \\
\textbf{COFT-AD} &88.36 &86.50 &85.87 &80.72 \\
\bottomrule
\end{tabular}
\end{table}

\subsection{Exploration on the Number of Augmentation $N_A$ for Anomaly Scoring}

In this section, we compared the results under different $N_A$. The results in \ref{tab:NA} show that increasing N\_A leads to a certain degree of improvement in performance. However, it comes with an increase in computational complexity and time costs and the overall anomaly detection accuracy saturates after increasing the number of augmentations.

\begin{table}[!htp]\centering
\caption{Comparison between using different $N_A$. Experiments on Flower17 dataset, Daffodil (normal) / Tulip (abnormal).}\label{tab:NA}
\setlength\tabcolsep{3pt}
\resizebox{0.75\linewidth}{!}{
\begin{tabular}{lcccc}\toprule
$N_A$ & 1 & 5 & 20 & 50 \\\midrule
\textbf{COFT-AD~(w/o np)} & 66.93 & 68.00 & 69.39  & 70.01\\
\textbf{COFT-AD~(w/ np)} & 70.13 & 74.13 & 76.15  & 77.70 \\
\bottomrule
\end{tabular}
}
\end{table}

\section{Conclusion}
Unsupervised anomaly detection often requires training on large unlabeled normal samples. Such anomaly-free dataset is not always available before the inference stage and anomaly detection may have to be carried out with limited training samples.

To fill this gap, we proposed a few-shot anomaly detection approach through fine-tuning a model pretrained on large external image collections to few-shot normal samples from the target task. We achieve this fine-tuning by optimizing a contrastive loss and cross-instance positive pair loss. When prior knowledge on possible anomalies is available we further incorporate a negative pair loss to separate normal sample embeddings from the synthesized negative samples. We extensively evaluated the performance of the proposed method on three controlled anomaly detection datasets and four real industrial defect detection datasets. Our method achieved state-of-the-art performance on all datasets when only a handful of normal samples are available. Finally, we show that the benefit of using synthetic negative samples is task-dependent and should only be considered when accurate prior knowledge is available. We also notice that the proposed contrastive training method does not significantly benefit from more normal samples for industrial anomaly detection tasks due to the low intra-class variation.

\section*{Acknowledgment}
This research is supported by the Agency for Science, Technology and Research (A*STAR) under its AME Programmatic Funds (Grant No. A20H6b0151).

\ifCLASSOPTIONcaptionsoff
  \newpage
\fi

\bibliographystyle{IEEEtran}
\bibliography{reference}

\begin{IEEEbiography}
[{\includegraphics[width=1in,height=1.25in,clip,keepaspectratio]{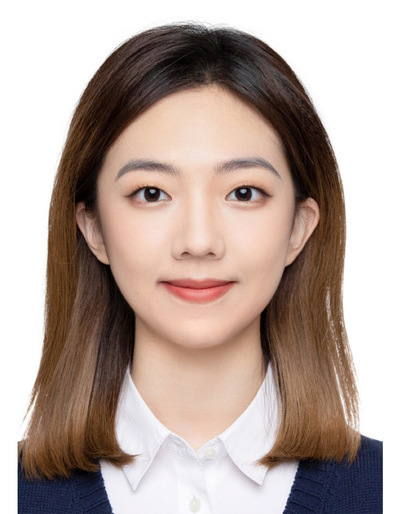}}]{Liao Jingyi}
Liao Jingyi received the B.E degree from Sichuan University, China, in 2018, and the Master degree from National University of Singapore in 2019. She is currently with the Institute for Infocomm Research (I2R), A*STAR, Singapore. Her research interests include anomaly detection, active learning and application in 3D data.
\end{IEEEbiography}

\begin{IEEEbiography}
[{\includegraphics[width=1in,height=1.25in,clip,keepaspectratio]{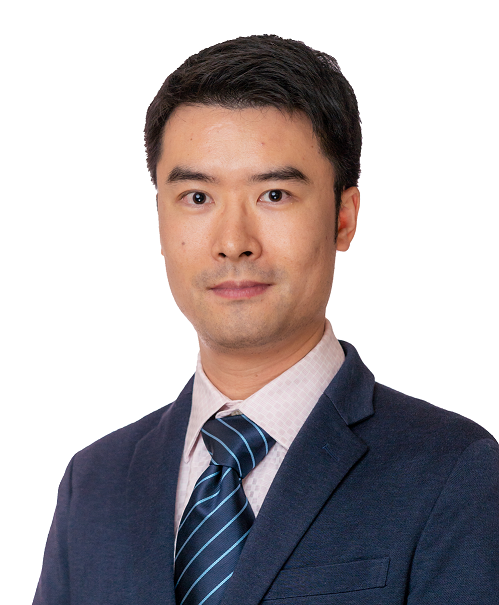}}]{Xu Xun}
Xun Xu (Senior Member, IEEE) received the B.Eng degree from Sichuan University in 2010 and the PhD degree from Queen Mary University of London in 2016. He was a research fellow with National University of Singapore between 2016 and 2019. He is now a senior scientist with I2R, A*STAR. His research interests include semi-supervised learning, domain adaptation, anomaly detection and zero-shot learning with applications to 3D point cloud data.
\end{IEEEbiography}

\begin{IEEEbiography}
[{\includegraphics[width=1in,height=1.25in,clip,keepaspectratio]{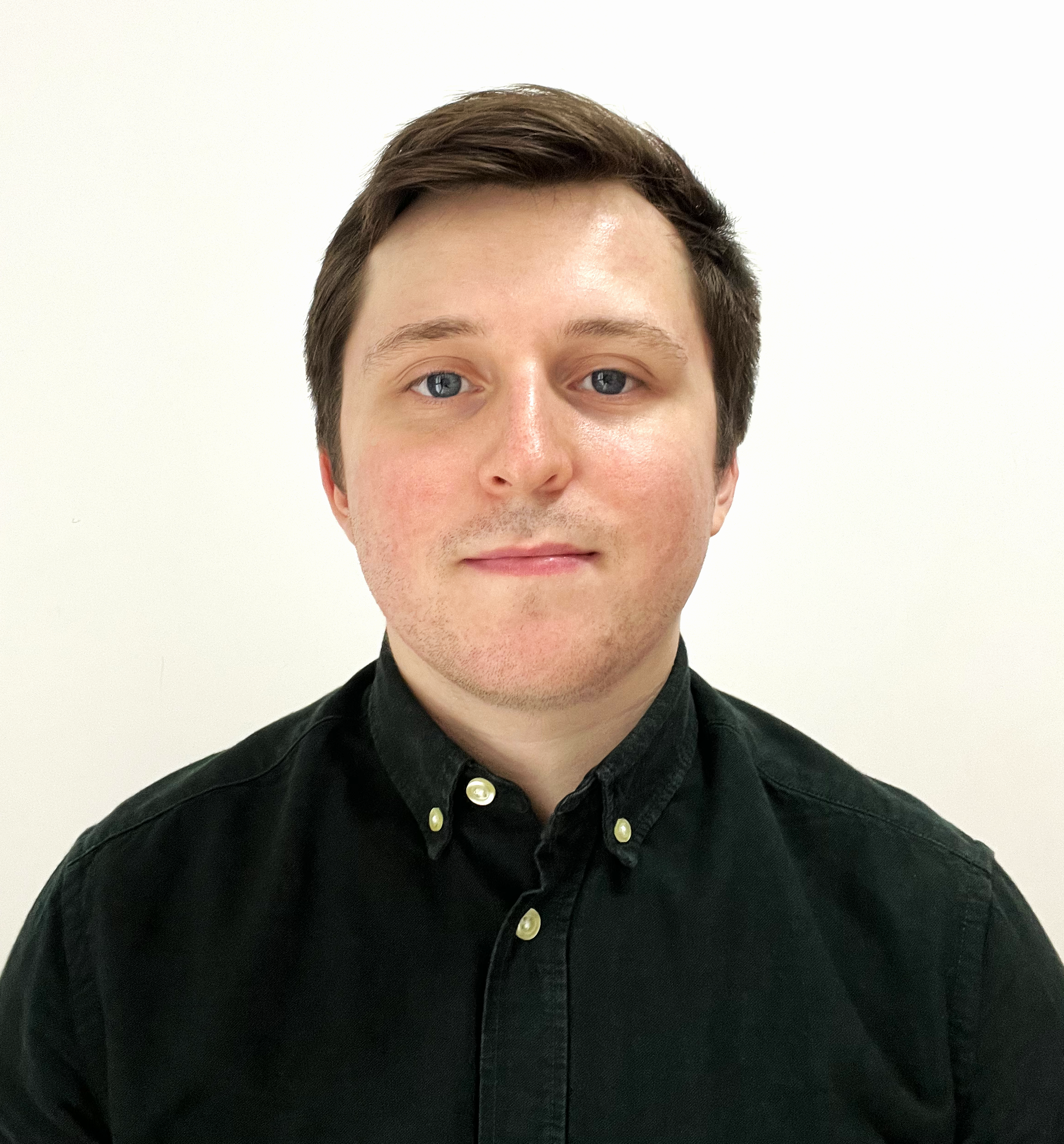}}]{Adam Goodge}
Adam Goodge is with the Institute for Infocomm Research (I2R), Agency for Science, Technology and Research (A*STAR), 1 Fusionopolis Way, \#20-10 Connexis, Singapore 138632.
\end{IEEEbiography}

\begin{IEEEbiography}
[{\includegraphics[width=1in,height=1.25in,clip,keepaspectratio]{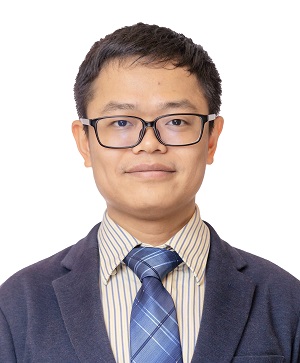}}]{Manh Cuong Nguyen}
Cuong Nguyen received the B.E. degree in computer engineering from Nanyang Technological University, Singapore, in 2014, and the Ph.D. degree in electrical and computer engineering from Carnegie Mellon University, Pittsburgh, PA, USA, in 2019. 
He is currently a scientist at Institute for Infocomm Research, A*STAR, Singapore. His research interests include semi-supervised learning, active learning, adversarial robustness, and their application in advanced manufacturing and medical imaging. 
\end{IEEEbiography}

\begin{IEEEbiography}
[{\includegraphics[width=1in,height=1.25in,clip,keepaspectratio]{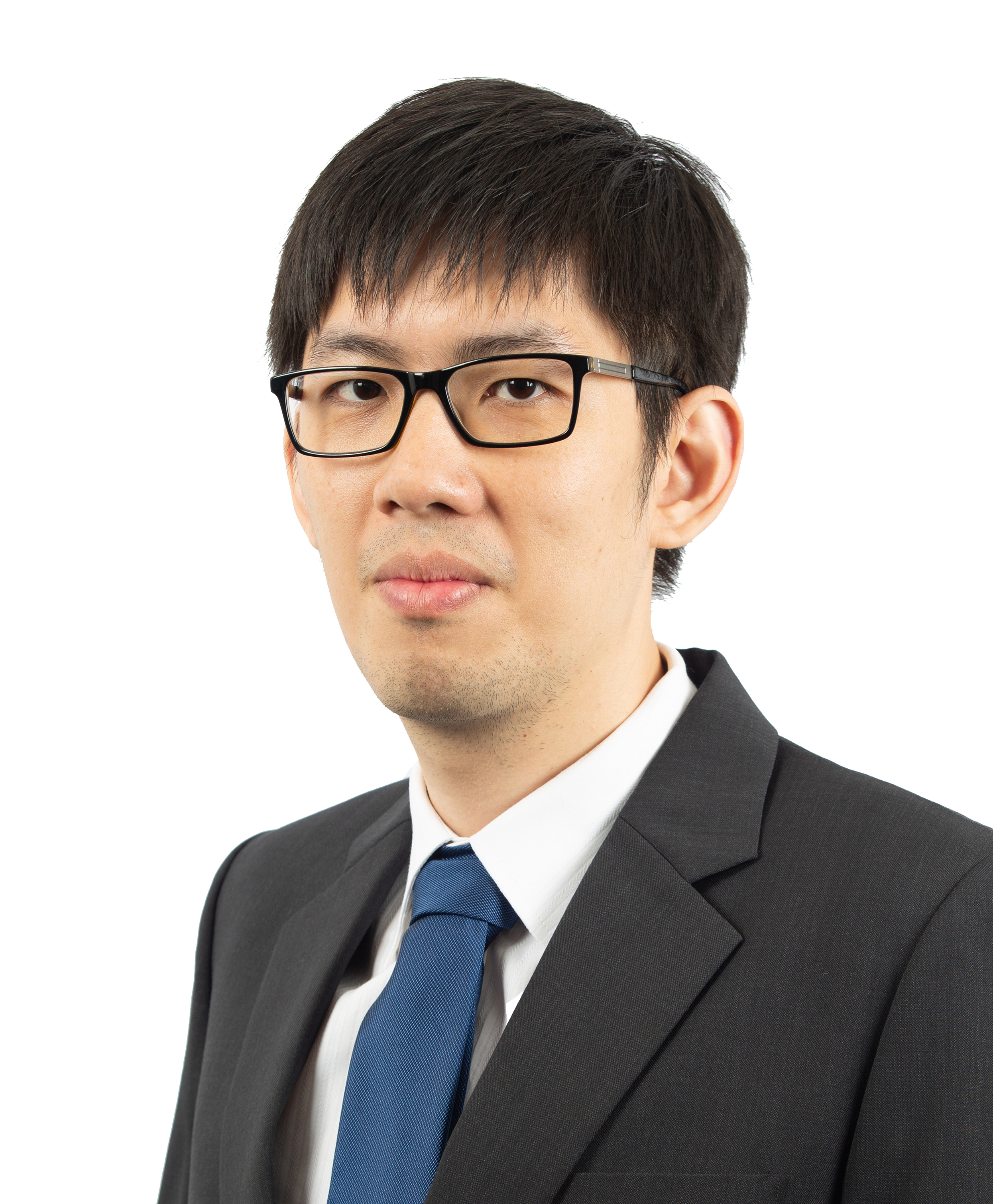}}]{Chuan Sheng Foo}
Chuan-Sheng Foo leads research in data-efficient and robust deep learning as Assistant Head of Department (Research), Machine Intellection Department, Institute for Infocomm Research (I2R) and as an Investigator at the Centre for Frontier AI Research (CFAR), A*STAR. He received BS, MS, and PhD degrees in Computer Science from Stanford University.
\end{IEEEbiography}
\end{document}